\documentclass{article} % For LaTeX2e
\usepackage{iclr2027_conference,times}
\usepackage[hidelinks]{hyperref}
\usepackage{url}
\usepackage{graphicx}
\usepackage{booktabs}
\usepackage{amsmath}
\usepackage{amssymb}
\usepackage{xcolor}
\usepackage{multirow}
\usepackage{wrapfig}
\usepackage{enumitem}

\title{Linger and Lose: Knowledge Collapse in Low-Bit Language Models}

\author{Prashanna Mani Paudel, Shivanand Venkanna Sheshappanavar\\ Geometric Intelligence Research Lab, \\Dept. of Electrical Engineering and Computer Science, University of Wyoming, USA.\\ \{ppaudel,ssheshap\}@uwyo.edu}
\iclrfinalcopy
\begin{document}

\maketitle

\begin{abstract}
Training language models with ternary weights is commonly judged by loss and downstream accuracy, which record only a modest cost relative to full precision. We show that these metrics can conceal a much larger failure. We instead measure knowledge capacity, the factual bits stored per parameter, on synthetic biographies with known information content. We train GPT-2-style models from scratch with 2.5M to 50M parameters at five precisions. Under the standard cosine schedule, ternary models retain as little as \textbf{6\%} of an identically trained fp16 model's capacity. The deficit widens with model size while perplexity rises by only \textbf{1.4} to \textbf{1.6} times. Measured throughout training, these models first acquire capacity and then lose most of it. We identify this knowledge collapse as a learning-rate dwell instability. Held near $1$--$2\times10^{-4}$ with no decay, a pre-collapse model collapses within a few hundred exposures, and returning to a safe rate does not restore capacity. We then locate the collapse in the output head. It happens at a value the model can never predict. The weights there grow unchecked, while every other prediction the model makes is unchanged. We find that making the value predictable removes the collapse, regardless of which attribute carries it. A warmup-stable-decay schedule with a 10\% cooldown increases ternary capacity by \textbf{2.7} times at 25M and \textbf{4.3} times at 50M. Gains are larger at lower precision. Cutting the output head's learning rate prevents failure when training from scratch. The post-training quantization methods we tested recover no measurable capacity below 4 bits. Our findings suggest judging low-precision training by retained capacity during training rather than final loss.
\end{abstract}

\section{Introduction}
\label{sec:intro}

Language models trained with ternary weights are usually judged by two metrics: how much the loss rises and how many benchmark points the model loses. Both can stay nearly unchanged while the model's measured knowledge capacity falls by more than half. \citet{allenzhu2024capacity} showed that loss is a poor proxy for stored knowledge. They measured knowledge directly by training on synthetic biographies of known informational content and calculating learned bits per parameter. Their measurement covered only post-training quantization (PTQ), and they conjectured that quantization-aware training (QAT) would behave differently. We test that conjecture for precisions of int4 and below. Loss-based results imply that capacity degrades gradually as precision falls; we find instead that ternary models acquire capacity and then lose most of it to instability during training.

We train GPT-2-style models from scratch at fp16, int8, int4, int3, and ternary precision, following \citet{bitnetb158}, at four sizes from 2.5M to 50M parameters. We measure \emph{retention}, the fraction of an identically trained fp16 model's capacity that it retains, and find that int8 retains more than 90\% of the fp16 capacity. Below int4, retention drops sharply, and under the standard cosine schedule, it falls further as the model grows from 0.36 at 2.5M to 0.06 at 50M. A representational limit would not predict this. As the number of people grows with the model, each person has the same parameter budget at every size, and a weight-precision limit would yield the same retention at every size. Previous work on ternary quantization evaluated through loss or accuracy ~\citep{spectra2024trilm,nielsenreloaded2024,paretoq2025} misses this because loss is insensitive to how much capacity the model loses.

The model size dependence is driven by the knowledge load, which is the number of people $N$ that the model must learn, and not by the parameter count. Keeping $N$ fixed while the model grows removes that dependence (\S\ref{sec:curve}). The collapse appears only when $N$ grows with the model. Measuring capacity throughout training shows the load effect. At full load, ternary models at 25M and 50M first acquire capacity, then lose most of it before training ends. The failure is therefore not a capacity ceiling on ternary weights but training instability. We show that this collapse is a learning-rate dwell instability, and that in our experiments, the schedule, not the precision, decides whether it occurs.

\textbf{Contributions:}
\begin{enumerate}[leftmargin=*]
\item We measure knowledge capacity under quantization-aware training below int4, which prior work did not, and the PTQ methods we tested recover no measurable capacity below 4 bits (\S\ref{sec:curve}, \S\ref{sec:ptq}).
\item We show that the size-dependent gap is a late-training capacity collapse driven by knowledge load rather than by parameter count. We identify it as a learning-rate dwell instability: a pre-collapse model held near $1$--$2\times10^{-4}$ collapses without decay and does not recover at a safe rate (\S\ref{sec:collapse}, \S\ref{sec:dwell}).
\item We show, with nine dataset variants, that the collapse requires a value the model never learns to predict and that it is indifferent to which attribute carries that value and where it sits (\S\ref{sec:mech}).
\item We show that a warmup-stable-decay (WSD) schedule with a 10\% cooldown removes the size-dependent decline in ternary retention and that slowing the output-head learning rate prevents the collapse from scratch (\S\ref{sec:fix}, \S\ref{sec:mech}).
\end{enumerate}

\section{Related Work}
\label{sec:related}

\textbf{Knowledge capacity:} Our capacity measurement follows \citet{allenzhu2024capacity}, who measured roughly 2 bits per parameter for GPT-2-style models on synthetic biographies. They showed a large drop under post-training GPTQ: int8 preserves capacity, while int4 loses more than half. They conjectured that QAT may be necessary for high-quality int4 models. We measure capacity under QAT itself, at int4 and below, and find that the deficit forms during training. A ternary weight can store at most $\log_2 3 \approx 1.58$ bits, and none of the capacity we report approaches that value. \citet{morris2025memorization} provides a quantization-free capacity estimator, which we use as a cross-check (Appendix~\ref{app:morris}). \citet{giorlandino2026} derives capacity thresholds for associative memories with continuous weights, and \citet{gardner1988space,gardnerderrida1988,krauthmezard1989} showed that restricting weights to a few discrete values lowers the maximum. Both results describe how much a model could store at most; our question is how much of what it has stored survives training.

\textbf{Low-bit training and its evaluation:} Prior work on low-bit training is judged by loss or accuracy. \citet{kumar2024precision} model quantized-training loss as a function of effective parameter count, and their fit predicts retention near 0.25 at 1.58 bits, close to our measurement under a short cooldown. That value needs a bias term they report only in an appendix, and 1.58 bits extrapolates below the 3 to 16 bits at which they pretrain. \citet{bitnetb158} popularized ternary pretraining, and Spectra~\citep{spectra2024trilm}, \citet{nielsenreloaded2024} and ParetoQ~\citep{paretoq2025} evaluate it through loss or accuracy, reaching different verdicts on its cost. Our measurement observes a collapse that neither detects, which may explain the disagreement. \citet{dettmers2023case} found the total-bits-versus-accuracy trade-off turning against 3-bit inference, and our capacity measurement also turns. \citet{saequant2026} show that perplexity can improve under quantization while interpretable features degrade, the same disagreement between loss and internal state that motivates our measurement.

\textbf{Schedules and optimizers in low-bit training:} Existing work has studied each schedule and optimizer choice we vary, but through loss. \citet{sub100mqat2026} finds that 33\% of the loss-optimal cooldown is across fp16, int8, and int6, and at int4, the choice is reported to be noise-dominated below 50M, our headline size. We test that fraction and find that loss and capacity disagree (\S\ref{sec:fix}). \citet{bitbybit2026} remove a convergence instability in 2-bit QAT with a progressive-precision curriculum, which we run against a control (\S\ref{sec:fix}). \citet{nagel2022oscillations} show latent weights oscillating across quantizer thresholds, driven by the straight-through estimator's gradient rather than by the learning rate. Our failure sits in the output head rather than the body and grows monotonically rather than oscillating. \citet{muon2026} reports Muon ~\citep{jordan2024muon} outperforming Adam on tail-end associative memory; we use it only to locate where the collapse can be controlled and find no capacity advantage under a short cooldown (\S\ref{sec:mech}). A collapsed model that does not recover at a safe rate is closest to the loss of plasticity ~\citep{dohare2024plasticity}.

\textbf{Late learning-rate dynamics and forgetting:} Three recent papers precede us on the schedule. \citet{catalantatjer2026ptqdynamics} show that post-training quantization error rises sharply once the learning rate decays, even as validation loss continues to improve. \citet{watts2026sharpness} show that a shorter anneal, a higher peak rate, or sharpness-aware pretraining each leave a model less affected by later fine-tuning or 4-bit quantization. \citet{rofin2026overtraining} tie decay-induced sharpening to forgetting under fine-tuning. All three pretrain in full precision and measure what survives a perturbation applied afterward. We add the same schedule dependence inside low-bit training, where capacity is lost while the model is still learning. \citet{jagielski2023measuring} found that memorized examples are forgotten the fastest just after a learning-rate decay step, at the post-decay rate, which depends on dwell time at the lower rate. In low-bit training, the same dependence appears as an abrupt collapse concentrated on one attribute rather than as gradual forgetting.

\section{Setup}
\label{sec:setup}

We vary precision, model size, and the learning-rate schedule, keeping the rest of the procedure the same (Appendix~\ref{app:details}).

\textbf{Task and metric:} We follow the bioS protocol of \citet{allenzhu2024capacity}. Each of $N$ synthetic people has six attributes (birth date, birth city, university, major, employer, gender) drawn from pools of known size and written out as short biographies. Birth date is composite, written as a month, day, and year drawn from 12, 28, and 200 values. Because we know the data's information content, we can count the bits a model has learned from its loss. With $p_1$ denoting the loss on a person's name and $p_2$ denoting the summed loss on the six attributes, both measured in nats, the \emph{capacity} is
\begin{equation}
R = \frac{N \cdot \big(\max(0,\, \log_2 N_0 - p_1 \log_2 e) + \max(0,\, \log_2 S_0 - p_2 \log_2 e)\big)}{P},
\end{equation}
in bits per parameter. Here $N_0$ is the number of possible names, $S_0$ the number of possible attribute combinations, and $P$ the parameter count. \emph{Retention} is $R_{\text{quant}}/R_{\text{fp16}}$, where the fp16 model has the same size, data, and exposures, and always runs the standard cosine schedule. So changes in retention across schedules come from the quantized model alone. The $\max(0,\cdot)$ affects only collapsed cells, which we mark $\dagger$ and report under two other estimators (Appendix~\ref{app:convention}).

\textbf{Models and knowledge load:} We train GPT-2-style decoders with rotary embeddings from scratch at four sizes (2.5M to 50M parameters) and five precisions: fp16, int8, int4, int3, and ternary ~\citep{bitnetb158}. We call the transformer blocks the \emph{body}. The embedding and the output head are quantized to int4 in every low-bit run, so ``ternary'' means a ternary body with an int4 embedding and output head. Only weights are quantized; activations stay in bf16. Following \citet{allenzhu2024capacity}, we give each model more knowledge than it can store. We set $N$ so the data hold 2.25 bits per parameter at every precision; we call this \emph{full load}. \emph{Light load} keeps $N$ fixed at 371k while the model size increases (\S\ref{sec:curve}); Appendix~\ref{app:details} lists the sizes and loads.

\textbf{Training and schedules:} We train with AdamW at a peak learning rate of $5\times10^{-4}$ for 1000 \emph{exposures}, where one exposure presents every person's facts once. \emph{Cosine} decays from the peak to 10\% of it; the low-bit work we compare against uses it, so we call it the standard schedule. \emph{Constant} holds the peak rate throughout. \emph{WSD} ~\citep{hu2024minicpm,hagele2024scaling} holds the peak rate and then applies a cosine cooldown over the final fraction $f$ of training, with $f=0.1$ unless stated.
 
\textbf{Evaluation and the collapse criterion:} We evaluate capacity on a fixed subsample every 25 exposures and on the full corpus at the end of training, which gives the final $R$. The capacity of a collapsed run \emph{flickers} between low and partial levels across evaluations, so we define collapse over two of them. A run has \emph{collapsed} when $R$ falls below $0.2$ on two consecutive evaluations after exceeding $0.25$, with the first being its \emph{onset}. A decline that does not meet this criterion is an \emph{erosion}. Every collapse count in the paper uses this criterion, and each table states its seed counts.

\section{The Capacity Gap That Loss Cannot See}
\label{sec:curve}

\begin{table}[t]
\caption{Ternary retention falls with model size while the loss penalty stays small. \textbf{Left}: retention $R_{\text{quant}}/R_{\text{fp16}}$ under the standard cosine protocol; $\dagger$: depends on the clamp convention (Table~\ref{tab:conv}). \textbf{Right}: the perplexity penalty ternary pays relative to fp16 against its capacity gap; PPL$\times = e^{\Delta \text{loss}}$, capacity gap $= 1/\text{retention}$. Appendix~\ref{app:details} gives the seed count of every cell.}
\label{tab:grid}
\centering
\small
\begin{minipage}[t]{0.5\linewidth}
\centering
\begin{tabular}[t]{lcccc}
\toprule
retention ($\uparrow$) & 2.5M & 10M & 25M & 50M \\
\midrule
int8 & 0.979 & 0.936 & 0.926 & 0.754 \\
int4 & 0.893 & 0.686 & 0.581 & 0.316$^\dagger$ \\
int3 & 0.660 & 0.468 & 0.320 & 0.504 \\
ternary & 0.364 & 0.235 & 0.090$^\dagger$ & 0.058$^\dagger$ \\
\bottomrule
\end{tabular}
\end{minipage}\hfill
\begin{minipage}[t]{0.47\linewidth}
\centering
\begin{tabular}[t]{lccc}
\toprule
size & $\Delta$loss ($\downarrow$) & PPL$\times$ ($\downarrow$) & gap ($\downarrow$) \\
\midrule
2.5M & +0.33 & 1.39$\times$ & 2.7$\times$ \\
10M & +0.34 & 1.41$\times$ & 4.3$\times$ \\
25M & +0.46 & 1.58$\times$ & 11.1$\times$ \\
50M & +1.06 & 2.88$\times$ & 17.1$\times$ \\
\bottomrule
\end{tabular}
\end{minipage}
\end{table}

Table~\ref{tab:grid} shows retention under the standard cosine schedule. Below int4, the losses are large, and for ternary, they deepen at every size, from 0.364 at 2.5M to 0.058 at 50M. The knowledge load grows with the model, so each person has the same parameter budget at every size, and yet the larger ternary models keep a smaller share of what fp16 stores (Figure~\ref{fig:retcurve}, Appendix~\ref{app:convention}).

\textbf{Loss understates the gap:} Table~\ref{tab:grid} (right) compares the perplexity penalty of ternary relative to fp16 with the capacity gap at the same size. From 2.5M to 25M, the penalty moves from $1.39\times$ to $1.58\times$, while the capacity gap grows fourfold, from $2.7\times$ to $11.1\times$. At 50M, two of three ternary seeds end with a training loss above 2, and the penalty reaches $2.88\times$ against a $17.1\times$ gap. Loss therefore captures only a small part of a large and growing gap, and evaluations of ternary training that rest on loss inherit that blindness. A depth-versus-width control and five changes to the training recipe peak at the same capacity, $R=0.35$--$0.40$, so none of the changes we tested removes the gap (Appendix~\ref{app:controls}).

\begin{wrapfigure}{r}{0.41\linewidth}
\vspace{-\baselineskip}
\centering
\includegraphics[width=\linewidth]{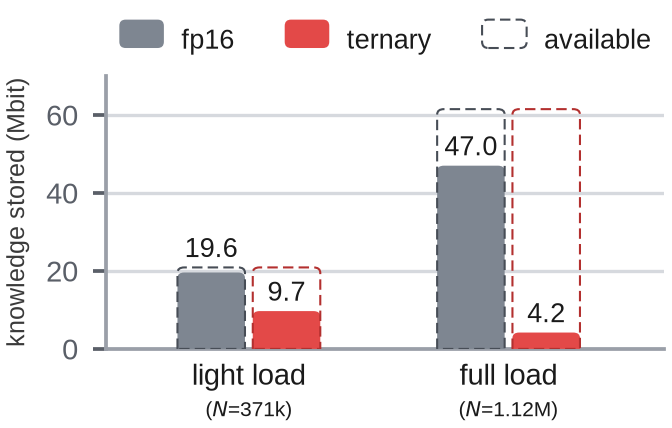}
\caption{When the knowledge load triples, fp16 stores more and ternary stores less. Learned megabits, $R\cdot P$, at 25M (mean over seeds).}
\label{fig:retention}
\vspace{2pt}
\end{wrapfigure}

\textbf{The gap follows the knowledge load, not the model size:} At full load, model size and knowledge load grow together, so Table~\ref{tab:grid} cannot separate them. A control does. We train 25M models at light and full load and change nothing else. The fp16 model stores more at full load than at light load; the ternary model stores \emph{less}, although full load has three times as many people (Figure~\ref{fig:retention}). At light load, ternary retention is 0.49 at 25M and 0.78 at 50M, compared to 0.090 and 0.058 at full load (Appendix~\ref{app:controls}). Full load also takes three times as many optimizer steps as light load, so we train the light load 25M model for 3029 exposures to match the step count. It does not collapse, and its capacity rises to $R=0.392$ and $0.385$ in two seeds. The number of steps does not explain the gap; the knowledge load does. The retention curve therefore measures what happens when a ternary model must store more than it can retain. The next section traces capacity over training and locates where it is lost.

\section{The Gap Is a Late-Training Collapse}
\label{sec:collapse}

\begin{figure}[t]
\centering
\includegraphics[width=\linewidth]{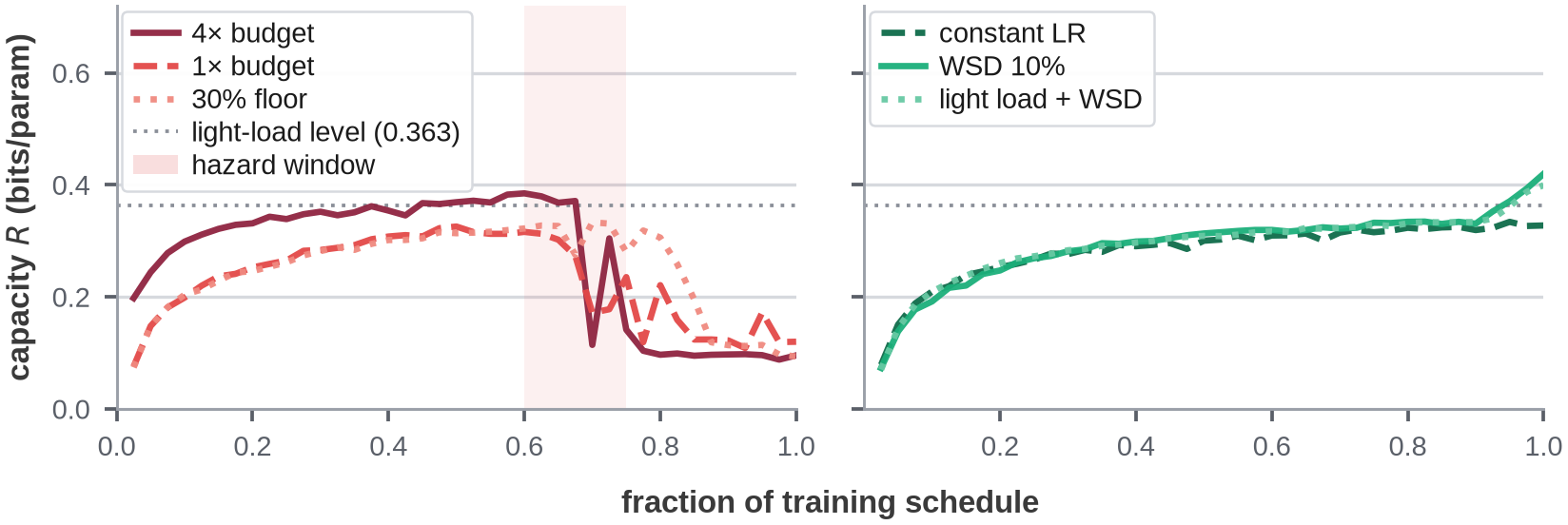}
\caption{Continuous capacity traces, 25M ternary at full load, one line per run, against the fraction of each run's schedule. \textbf{Left}: decaying schedules (cosine at $1\times$ and $4\times$ budget, cosine to a 30\% floor). \textbf{Right}: constant LR, WSD with a 10\% cooldown, and WSD at light load; the dotted line is the light-load reference.}
\label{fig:instability}
\end{figure}

A final checkpoint cannot distinguish a model that never acquired capacity from one that acquired it and lost it. So we measure capacity throughout training. Under the standard cosine schedule, full-load 25M ternary runs rise toward the light-load level, peak, and then lose most of their capacity (Figure~\ref{fig:instability}, left). We also trained two seeds for 4000 exposures (4$\times$ longer) and saw the same collapse, so longer training does not prevent it. At 50M, the collapse comes earlier: two of three seeds collapse a third of the way through the schedule, and the third collapses late.

The collapse begins at the same fraction of the schedule, whether training lasts 1000 or 4000 exposures, so it follows the schedule and not the training time. Read as a learning rate, it begins between $1.0$ and $1.6\times10^{-4}$ in all five 25M configurations we traced (Appendix~\ref{app:onset}). Halving the batch size leaves that rate unchanged, which a gradient-noise account does not predict. The 50M seeds that collapse early do so at $3.8\times10^{-4}$, so the band belongs to the 25M model and is not constant across sizes. Holding the learning rate at its peak removes the collapse in all four constant-rate runs (Figure~\ref{fig:instability}, right). These runs cannot tell whether capacity is lost by decaying through this range or by dwelling in it; \S\ref{sec:dwell} separates the two.

\textbf{The collapse is abrupt and narrowly localized:} Birth date, the one compositional attribute, swings between storing several bits per person and being anti-predictive from one evaluation to the next. Birth city, university, and major stay steady across the same evaluations (Figure~\ref{fig:bistab}). Pooled over nine crashing runs, birth date stays above half its peak in 25--33\% of post-peak checkpoints, and the three steady attributes in nearly all (Appendix~\ref{app:attributes}). Employer and gender match birth date's rate on a near-zero quantity, so survival does not order by cardinality.

\textbf{The same failure appears at full precision:} One of three 50M fp16 seeds crashes at 70\% of the schedule. Its \emph{name-prediction} loss rises from 16 to 70 nats (Figure~\ref{fig:bistab}, right, Appendix~\ref{app:attributes}). Low precision, therefore, appears to lower the threshold of a failure it did not invent. Within a run, the training loss does rise at the crash by $0.14$--$0.19$ nats, but the rise is small compared with the capacity lost. Loss misleads chiefly when you compare final losses across models, as Table~\ref{tab:grid} does.

\textbf{Where the increase sits:} Scoring 16 archived 25M checkpoints per token position locates it precisely (Appendix~\ref{app:audit}). In all 11 collapsed checkpoints, the four steady attributes—the name and the birth day and year—sit within noise of their pre-collapse parents. The whole increase is on the birth \emph{month}, which every ternary checkpoint predicts at chance, as does fp16 at both sizes; \S\ref{sec:mech} traces it to the output head.

\section{A Dwell-Dependent Hazard Window}
\label{sec:dwell}

To distinguish collapse from decay at a learning rate, we branch the pre-collapse final checkpoints of two constant-LR runs to a fixed learning rate and hold them there. The parents end at $R=0.338$ and $0.353$ after 1000 exposures, and each branch keeps its parent's optimizer state. We test nine learning rates from $5\times10^{-4}$ down to $0.5\times10^{-4}$, two seeds at each, for 500 exposures (250 at the top rate); Table~\ref{tab:dwell} lists the endpoints, and Appendix~\ref{app:dwell} lists the traces. Within a seed, every run starts from the same pre-collapse model, so the only variable is where it is parked.

\begin{table}[t]
\caption{Held-learning-rate dwell map, 25M ternary full load, branched from the pre-collapse constant-LR checkpoints: final $R$ per seed and, for collapses under the \S\ref{sec:setup} criterion, onset in exposures of dwell; hold length in parentheses. $\dagger$: depends on the clamp convention (Table~\ref{tab:conv}).}
\label{tab:dwell}
\centering
\footnotesize
\setlength{\tabcolsep}{4pt}
\begin{tabular}{llll}
\toprule
held LR (hold) & final $R$ (s0 / s1, $\uparrow$) & onset (s0 / s1) & outcome \\
\midrule
$5.0\times10^{-4}$ (250) & 0.348 / 0.360 & -- & flat (control) \\
$4.0\times10^{-4}$ (500) & 0.355 / 0.368 & -- & stable \\
$3.0\times10^{-4}$ (500) & 0.362 / 0.107$^\dagger$ & -- / $470$ & s1 late collapse \\
$2.0\times10^{-4}$ (500) & 0.101$^\dagger$ / 0.191$^\dagger$ & $220$ / $410$ & both collapse; s1 flickers, then collapses \\
$1.5\times10^{-4}$ (500) & 0.236 / 0.109$^\dagger$ & $260$ / $170$ & both collapse; s0 partial recovery at the end \\
$1.2\times10^{-4}$ (500) & 0.105$^\dagger$ / 0.182$^\dagger$ & $270$ / $290$ & erosion to 250, then both collapse \\
$1.0\times10^{-4}$ (500) & 0.110$^\dagger$ / 0.158$^\dagger$ & $250$ / $270$ & both collapse \\
$0.8\times10^{-4}$ (500) & 0.111$^\dagger$ / 0.195$^\dagger$ & ${\approx}420$ / $360$ & late collapse \\
$0.5\times10^{-4}$ (500) & 0.215 / 0.188$^\dagger$ & -- / $430$ & s0 heavy flicker, eroded; s1 late collapse \\
\bottomrule
\end{tabular}
\end{table}

\textbf{Collapse needs no decay, only time at a dangerous learning rate:} A pre-collapse model parked in the $1$--$2\times10^{-4}$ band with no decay at all collapses in all eight runs we held there. Onsets fall between 170 and 410 exposures, and nothing at or below $2\times10^{-4}$ holds its starting capacity. LR decay doesn't cause the collapse; time spent at these rates does.

\textbf{Outside the band, the hazard weakens but does not stop:} Safety at 250 exposures does not mean safety at 500. Of the eight arms we held outside the band for the full 500 exposures, four collapsed, the earliest at exposure 360. At $3\times10^{-4}$, a mild erosion by 250 exposures becomes a collapse in one of two seeds by 500. By $4\times10^{-4}$, the hazard is gone at this horizon, and the constant-LR arms hold the peak rate for the whole run. The map, therefore, measures a hazard \emph{rate} that varies continuously with the learning rate and peaks near $1$--$2\times10^{-4}$ (Figure~\ref{fig:hazard}). A fixed-length hold censors its tails, so the width we read off the map depends on how long we hold. Muon on the body, with AdamW on the embedding and output head, sees the same window. The onset band under cosine therefore reflects both how hazardous each learning rate is and how long the schedule spends there.

\textbf{Seeds change the timing, not the outcome:} At the same held rate, seeds collapse at different times, one at 220 exposures and the other at 410 under $2\times10^{-4}$. Recovery is transient: of the 13 held-rate arms that fall below $0.2$ after exceeding $0.25$, 12 climb back above $0.25$ at least once, and none ends the hold there (Appendix~\ref{app:dwell}). Collapsed traces do not settle but jump by a factor of three between evaluations, so we report onset under an operational definition rather than endpoints.

\textbf{A safe rate does not undo the collapse:} We returned three collapsed checkpoints to the safe constant rate for 250 exposures. They average $0.13$ to $0.16$ over their last 50 exposures, well under their pre-collapse level. A \emph{fresh} model at that rate reaches $0.27$ in the same budget. Resetting the optimizer state, wholly or for either parameter group, separates no arm from replicate noise (Appendix~\ref{app:dwell}). A collapsed model, therefore, does not regain its pre-collapse capacity at the rate that built it, a low-bit instance of loss of plasticity ~\citep{dohare2024plasticity}. We bound that to the 250 exposures we tested, and our results suggest avoiding the collapse rather than repairing it.

\textbf{Any drop gives an immediate gain:} We also found that every run that drops below $3\times10^{-4}$, including those that later collapse, first jumps from about $0.34$ to $0.40$--$0.49$ within ten exposures and then erodes or collapses over the following hundreds. Lowering the learning rate raises capacity at once, and staying at the low rate then loses it slowly.

\section{Dwell Time Predicts Every Schedule's Outcome}
\label{sec:fix}

\begin{figure}[t]
\centering
\includegraphics[width=\linewidth]{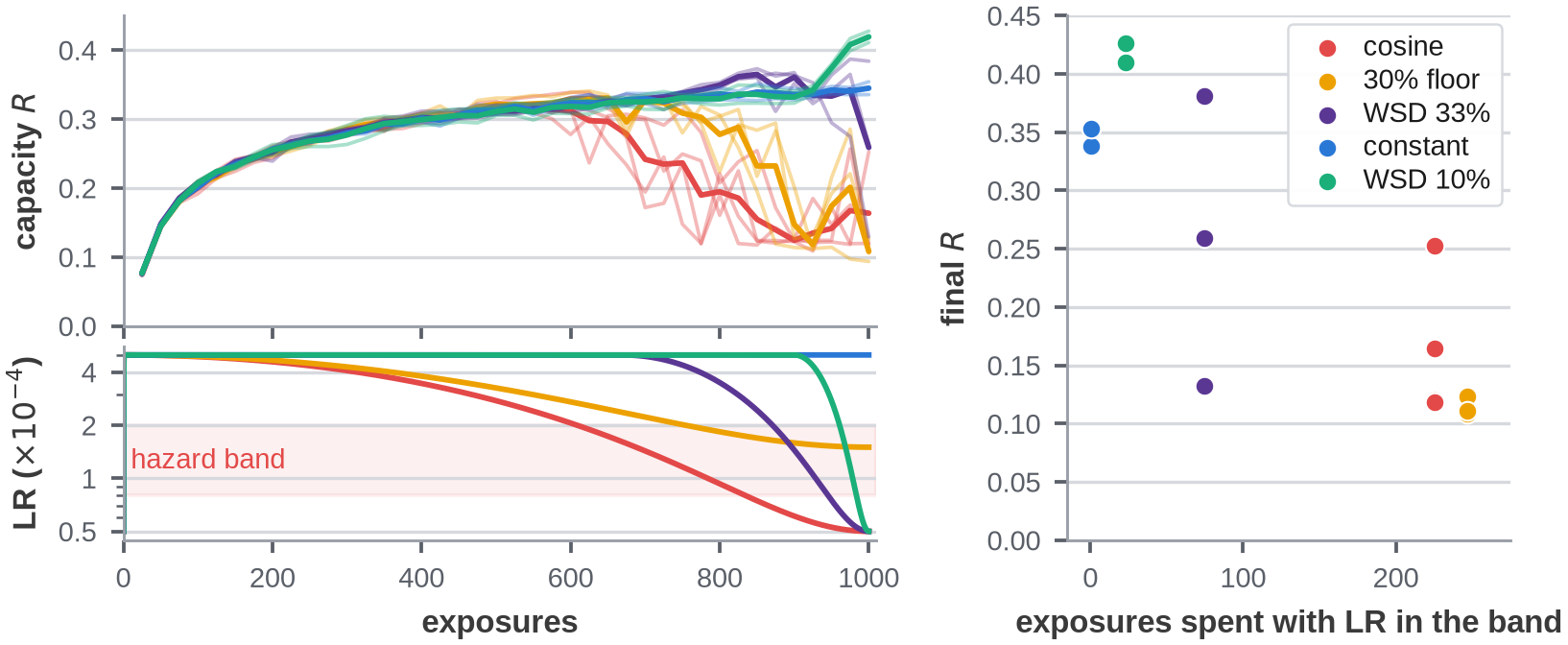}
\caption{\textbf{Left}: 25M ternary full-load traces (thin: seeds; bold: mean) under five schedules, each schedule's learning rate against the hazard window below. \textbf{Right}: final $R$ against exposures spent in the window.}
\label{fig:band}
\end{figure}

\begin{wraptable}{r}{0.48\linewidth}
\vspace{-\baselineskip}
\centering\footnotesize\setlength{\tabcolsep}{3.5pt}
\caption{Schedule gain by precision at 25M, full load: capacity under standard cosine vs WSD with a 10\% cooldown, and the relative gain.}
\label{tab:precgain}
\begin{tabular}{lccc}
\toprule
precision & cosine $R$ ($\uparrow$) & WSD $R$ ($\uparrow$) & gain \\
\midrule
fp16 & 1.720 ($n{=}6$) & 1.644 ($n{=}3$) & $-4\%$ \\
int8 & 1.593 ($n{=}2$) & 1.684 ($n{=}2$) & $+6\%$ \\
int4 & 0.999 ($n{=}2$) & 1.229 ($n{=}2$) & $+23\%$ \\
int3 & 0.551 ($n{=}2$) & 0.898 ($n{=}3$) & $+63\%$ \\
ternary & 0.155 ($n{=}6$) & 0.412 ($n{=}5$) & $+166\%$ \\
\bottomrule
\end{tabular}
\vspace{4pt}
\end{wraptable}
Section~\ref{sec:dwell} showed two effects of the learning rate, which we call the dwell account: dropping the rate below the window helps at once, and dwelling inside the window harms slowly. Every schedule that ends low receives the fast benefit, so schedules should differ by the slow harm, that is, by how long each dwells in the window. We count an exposure as dwell when the logged learning rate lies between $0.8$ and $2\times10^{-4}$, the range over which both seeds collapse at every rate we held (Table~\ref{tab:dwell}). Standard cosine dwells for about 230 exposures, the 30\%-floor variant for about 250, a 33\% cooldown for about 75, and a 10\% cooldown for about 23. Figure~\ref{fig:band} overlays the schedules on the window and plots final $R$ against this count. Constant LR is the one exception: it never dwells, but it also never drops, so it misses the fast benefit and finishes below the 10\% cooldown.

\textbf{A short WSD cooldown removes the scale-dependent decline:} A 10\% cooldown holds peak LR for 90\% of training and crosses the hazard window with an order of magnitude less dwell than cosine. Retention improves at every size we tested, and the decline in Table~\ref{tab:grid} disappears (Table~\ref{tab:wsd}). The aggregate clamp gives the largest gain, $2.7\times$ at 25M and $4.3\times$ at 50M, and the compression estimator of \citet{morris2025memorization} the smallest, $1.2\times$ at 25M (Table~\ref{tab:conv}). We therefore treat the magnitude as metric-dependent, though all three estimators place cosine below the 10\% cooldown at 25M and 50M. Two controls rule out simpler explanations. Under matched WSD, light and full load retain the same capacity, so the load penalty of \S\ref{sec:curve} is itself schedule-mediated. fp16 scores slightly lower under WSD than under cosine, so the gain is not a denominator artifact.

\begin{table}[t]
\caption{Ternary retention by size, standard cosine versus WSD. Indented row: the cosine cells under the per-attribute floor; $\dagger$: depends on the clamp convention (Table~\ref{tab:conv}).}
\label{tab:wsd}
\centering
\small
\begin{tabular}{lccc}
\toprule
schedule (retention) & 10M ($\uparrow$) & 25M ($\uparrow$) & 50M ($\uparrow$) \\
\midrule
standard cosine & 0.235 ($n{=}2$) & 0.090 ($n{=}6$) & 0.058 ($n{=}3$) \\
\quad per-attribute floor & 0.235 & 0.126 & 0.109 \\
WSD, 10\% cooldown & 0.257 ($n{=}3$) & 0.240 ($n{=}5$) & 0.247 ($n{=}3$) \\
WSD, 33\% cooldown & 0.241 ($n{=}2$) & 0.150$^\dagger$ ($n{=}3$) & 0.067$^\dagger$ ($n{=}1$) \\
\bottomrule
\end{tabular}
\end{table}

\textbf{The schedule gain is a low-precision effect, and it scales with bit-width:} int3 never meets the collapse criterion, yet a short cooldown lifts it at each of the three sizes we ran. Running the whole precision row at 25M under both schedules (Table~\ref{tab:precgain}) shows two regimes on one curve. Below int4, the gain is recovery from the collapse. At int8 and int4, there is no collapse to recover, yet the gain is still positive: the milder cost a long decay imposes on a low-bit model, even when nothing dramatic is visible. We therefore read the schedule as acting on a single mechanism whose severity is set by bit-width, and all three fp16 WSD runs fall below the cosine mean.

\textbf{Three tests of the dwell account:} Section~\ref{sec:dwell} implies that the time spent in the hazard window determines where a schedule ends up. We check this on three schedules.
\begin{enumerate}[leftmargin=1.5em,itemsep=0pt,topsep=1pt,parsep=0pt,partopsep=0pt,label=(\roman*)]
\item \emph{A 33\% cooldown} lands between the 10\% cooldown and cosine, as its dwell predicts (Table~\ref{tab:wsd}). \citet{sub100mqat2026} identifies 33\% as loss-optimal, so choosing a cooldown by loss does not account for knowledge; the divergence \citet{watts2026sharpness} reports at full precision. One of three seeds collapses, so we claim the ordering rather than a dose–response.
\item \emph{A 30\% floor} decays into the window and stays there. Two of three seeds collapse, and the third reaches the same final capacity, while two hold a pre-collapse level until late. What matters is accumulated exposure in the window, not whether the schedule is still decaying.
\item \emph{A precision curriculum}, fp16$\to$int8$\to$int4$\to$ternary with a re-warm at each stage, beats cosine but stays below WSD. A ternary model trained from scratch on the \emph{identical} staged learning-rate path reproduces the plain-cosine result (Figure~\ref{fig:curriculum}), so the benefit is the precision history, not the schedule. We take the recipe from \citet{bitbybit2026} and choose the stage fractions ourselves.
\end{enumerate}
Our results, therefore, favor the shorter of the two cooldowns we tested: 10\% and 33\%.

\section{Where Is the Control Point? Mechanism Triage}
\label{sec:mech}

Section \ref{sec:dwell} established when the collapse happens, not where it happens in the model or what is broken. This section asks both, and reports what we ruled out alongside what survived.

\textbf{The quantizer's own statistics show nothing at onset:} We logged how many body weights sit in each ternary state, how often they change state, and how large a typical update is against the quantizer's level spacing (Appendix~\ref{app:telemetry}). None of them move at onset in any crashing run, and a WSD cooldown ends below the onset value of all three without collapsing. Because these statistics are aggregated across layers, they rule out a model-wide quantizer signature at collapse onset, but not a change confined to a particular layer.

\textbf{The output head diverges at the token the model never learns:} The output head has one row per vocabulary token, and a row's size bounds its logit. In one collapsing cosine run, the mean norm of the 12-month rows grows from 18.5 to 41.7, while the mean of the rest falls from 2.6 to 2.2 under weight decay. The growth begins at exposure 525, as the schedule enters the hazard window, and the largest month logit rises from 431 to 11{,}879 (Appendix~\ref{app:audit}). Month accuracy remains at chance throughout, so the growth reflects neither learning nor month loss.

\textbf{What prevents the collapse also prevents the growth:} A 10\% cooldown keeps the logits in the hundreds at every precision, and cutting the output head's learning rate tenfold holds the month rows at $2.5$ against the pre-collapse parent's $18.6$. That cut also shrinks every other row of the head, so it limits the growth rather than selectively closing the gap between the month rows and the rest. We therefore read the collapse as unbounded growth of the head rows for a token whose gradient never vanishes (Appendix~\ref{app:audit}).

\textbf{The unlearned token causes the collapse, whichever attribute carries it:} Nine corpus variants test whether the collapse needs a token whose loss never falls, each changing one property of the data and nothing else (Table~\ref{tab:corpus}, Appendix~\ref{app:corpus}). Removing the birth date stops the collapse; a control that re-sizes $N$ by the same amount but keeps the date does not. Removing only the month stops it, and so does making it predictable from the first name, where the model then learns it. Moving the date to the end of the biography changes nothing, so position is not the cause. Widening the month from 12 values to 200 stops the collapse outright and gives the highest 25M ternary capacity we have recorded under cosine. A representational limit would predict the opposite. Deleting the birth date and injecting one unrelated 12-way tag returns both seeds to the collapsed level, with the growth now on the tag. No variant we ran collapses without such a token, and injecting one restores it. With two seeds each, we claim the direction of every effect rather than its size.

\begin{table}[t]
\caption{The collapse requires a token the model never learns, and is indifferent to which attribute carries it and where it sits. Corpus variants at 25M ternary under the standard cosine schedule, two seeds each; final $R$ per seed and onset in exposures under the \S\ref{sec:setup} criterion. $N$ is re-sized so that every variant holds 2.25 bits per parameter. Dashes: no onset; the arm never meets the criterion.}
\label{tab:corpus}
\centering
\footnotesize
\setlength{\tabcolsep}{5pt}
\begin{tabular}{llllc}
\toprule
variant & what it changes & final $R$ (s0 / s1, $\uparrow$) & onset & collapse \\
\midrule
unmodified & --- & 0.120 / 0.163 & 700 / 750 & yes \\
keep date & control for the re-sizing of $N$ & 0.127 / 0.138 & 525 / 550 & yes \\
no birth date & removes the unlearned token & 0.372 / 0.355 & -- & no \\
no month & removes the unlearned token only & 0.365 / 0.387 & -- & no \\
month from name & makes the month learnable & 0.376 / 0.387 & -- & no \\
date last & moves the date to the final sentence & 0.108 / 0.136 & 550 / 675 & yes \\
month 200-way & raises the month's payoff & \textbf{0.395 / 0.393} & -- & no \\
inject tag & no date, one random 12-way token & 0.130 / 0.131 & -- / 450 & s1 only \\
key--value corpus & no biography grammar at all & 0.370 / 0.365 & -- & no \\
\bottomrule
\end{tabular}
\end{table}

\textbf{The embedding and output head are sufficient control points:} Table~\ref{tab:opt} varies which parameters each optimizer updates under the crashing cosine schedule. Muon on the body only delays the decline, while Muon \emph{only} on the embedding and output head removes the collapse and rises to the best cosine result of any configuration. Muon is not required: cutting the same learning rate under plain AdamW reproduces the recovery, twentyfold or tenfold with $\eta\lambda$, the learning rate times the weight decay, held fixed. Under constant LR and WSD, both optimizers are stable, so placement is an instrument rather than an optimizer contest. Cutting the head's rate alone prevents the collapse, and cutting the embedding's alone does not. Leaving those two matrices at fp16 still collapses one of two cosine seeds (Appendix~\ref{app:controls}), so what is localized is their update, not their lattice.

\textbf{On a pre-collapse model, the output head buys time, not immunity:} We take pre-collapse checkpoints from the constant-LR runs and hold the body learning rate at $1\times10^{-4}$, where the baseline eventually collapses. We then test three variants, lowering the output-head learning rate, the embedding learning rate, or both tenfold, with weight decay compensated (Appendix~\ref{app:dwell}). On two lineages held for 300 exposures, the head and both-halves arms never fall below 0.28, while the embedding arm crosses 0.25 at 245 and 255. On a third lineage, held for 500 exposures against a pre-registered prediction, the head arm outlasts the baseline by over 100 exposures and never meets the criterion. From exposure 445, it shows the dips that precede every collapse we have recorded, and the prediction's birth-date clause failed. The ordering holds on three of three lineages; prevention holds on two within their horizons.

\section{The Limits of Post-Training Quantization}
\label{sec:ptq}

\begin{wraptable}{r}{0.45\linewidth}
\vspace{-\baselineskip}
\centering\footnotesize\setlength{\tabcolsep}{2.5pt}
\caption{Below int4 no post-training method recovers measurable capacity. Capacity $R$ on identical 25M checkpoints and lattice ($\uparrow$); 10M in Appendix~\ref{app:ptq}.}
\label{tab:ptq}
\begin{tabular}{lcccc}
\toprule
method & int8 & int4 & int3 & ternary \\
\midrule
RTN & 1.710 & 0.136 & 0.000 & 0.000 \\
AWQ & 1.672 & 0.122 & 0.000 & 0.000 \\
GPTQ & 1.674 & 0.142 & 0.000 & 0.000 \\
NF4 & -- & 0.174 & -- & -- \\
QAT, cosine (ours) & 1.593 & \textbf{0.999} & \textbf{0.551} & \textbf{0.155} \\
\bottomrule
\end{tabular}
\vspace{4pt}
\end{wraptable}

Alternatively, we can train at full precision and quantize afterward. We apply GPTQ ~\citep{gptq2022}, AWQ ~\citep{awq2023}, NF4 and round-to-nearest to our own fp16 checkpoints on the identical lattice and evaluator (Table~\ref{tab:ptq}; Appendix~\ref{app:ptq}). At int8, every method lands near fp16. Below int4, every method we ran scores exactly $0.000$, and fine-tuning a trained fp16 checkpoint into ternary converges to the from-scratch plateau (Appendix~\ref{app:controls}). For the methods we implemented and this corpus, knowledge below 4 bits has to be learned in low precision and protected while it is learned (\S\ref{sec:dwell}--\ref{sec:mech}). We did not test quantizers that learn their own codebook. \citet{catalantatjer2026ptqdynamics} obtain usable 3-bit GPTQ models on natural-language benchmarks, so what fails here is factual capacity rather than 3-bit quantization itself.

\section{Discussion and Conclusion}
\label{sec:discussion}

Extreme low-bit training is unstable in ways current recipes were not designed to avoid. A representational limit would call for architecture research; a stability failure calls for schedule and optimizer research. Our results suggest two inexpensive changes: a short cooldown and a smaller learning rate for the output head.

Our models top out at 50M parameters on synthetic biographies. Whether natural corpora contain values a model never learns is open, and we have not tested it. Even under WSD, ternary retention plateaus near a quarter of fp16, a residual gap we do not explain. On the one lineage we pre-registered, cutting the output head's rate delayed the collapse rather than preventing it. The dwell map is at 25M only, and we tested reversibility over a 250-exposure horizon. One question we leave open: a collapsed checkpoint's other predictions are unchanged, and a compression-based estimator reads no drop, so how much of the measured collapse is lost knowledge is unsettled.

Ternary models' reported robustness can reflect a metric blind to this failure rather than stable training. The collapse is a learning-rate dwell instability that a short cooldown removes. This work asks whether other low-bit failures are likewise schedule problems rather than capacity limits.

\subsection*{AI use statement:} We used large language models as coding assistants throughout the process of writing and debugging training, evaluation, and analysis code, as well as for polishing and editing the manuscript. They did not generate results, fabricate citations, or decide which findings to report. Every number traces back to a logged run or a deterministic script, and the corresponding author verified every claim and number against the raw logs and takes full responsibility.

\subsection*{Reproducibility statement:} We specify the capacity metric, lattice, schedules, optimizer placements, and collapse criterion in \S\ref{sec:setup}. All numbers come from logged training runs (final-checkpoint and continuous capacity traces) and deterministic post-hoc scripts applied to trained checkpoints. We record seeds and hyperparameters for every configuration in the run logs. We will release the code, logs, and analysis scripts with the camera-ready version.

\subsection*{Ethics statement:} This work uses only synthetic data. Each biography is procedurally generated and describes no real person, although the names, cities, and universities are drawn from pools of real ones. We train research-scale models of at most 50M parameters and release no deployed system.

\bibliography{references}
\bibliographystyle{iclr2027_conference}
\newpage

\appendix

% appendix floats: keep them on the page that references them
\renewcommand{\topfraction}{0.92}
\renewcommand{\bottomfraction}{0.75}
\renewcommand{\textfraction}{0.06}
\renewcommand{\floatpagefraction}{0.80}
\setcounter{topnumber}{3}
\setcounter{bottomnumber}{2}
\setcounter{totalnumber}{5}

\section{Training and Implementation Details}
\label{app:details}

\begin{table}[h]
\centering
\small
\caption{Hyperparameters shared by every run unless a section states otherwise.}
\label{tab:hparams}
\begin{tabular}{lp{0.62\textwidth}}
\toprule
optimizer & AdamW, $\beta_1=0.9$, $\beta_2=0.95$, decoupled weight decay 0.01 \\
gradient clipping & 1.0 \\
batch size & 192 \\
peak learning rate & $5\times10^{-4}$, after a 1000-step linear warmup \\
cosine floor & 10\% of the peak \\
training length & 1000 exposures \\
compute precision & bf16 \\
quantizer & groups of 64 weights; absmax scaling for integers, absmean for ternary ~\citep{bitnetb158} \\
gradient estimator & clipped straight-through ~\citep{yin2019ste}; varied in Appendix~\ref{app:ste} \\
embedding and output head & untied, int4 in every low-bit run \\
\bottomrule
\end{tabular}
\end{table}

\begin{wraptable}{r}{0.42\linewidth}
\vspace{-\baselineskip}
\centering
\small
\caption{Model sizes. The size is the label used throughout; $P$ is the built model's parameter count, which $R$ divides by; $N$ is the number of people at full load.}
\label{tab:sizes}
\begin{tabular}{lrr}
\toprule
size & $P$ & $N$ (full load) \\
\midrule
2.5M & 1.95M & 75k \\
10M & 9.29M & 371k \\
25M & 27.3M & 1.12M \\
50M & 47.1M & 1.96M \\
\bottomrule
\end{tabular}
\vspace{2pt}
\end{wraptable}

\textbf{Seed counts:} Table~\ref{tab:grid} uses six fp16 seeds per cell, three at 50M; two quantized seeds per cell, except int8 at 10M (four), ternary at 25M (six) and at 50M (three), and int8, int4, and int3 at 50M (one). The 50M fp16 reference is the mean of the two seeds that did not crash ($R=1.617$); including the third gives $1.536$ and a ternary retention of 0.061.

\textbf{Parameter count:} $P$ is the model's full trainable parameter count, including the embedding and output head. The size labels (2.5M to 50M) are the targets for the sizing rule, which chooses the model width so that the total count lands near the target. The width is rounded to a multiple of 64, so the built models differ from their labels (Table~\ref{tab:sizes}). Retention divides two values of $R$ at the same $P$, so it does not depend on this choice.

\textbf{The clamp:} Definition~4.3 of \citet{allenzhu2024capacity} imposes no non-negativity. The $\max(0,\cdot)$ is ours, and we apply it once to the aggregate value term. For a collapsed cell, a per-attribute floor reads higher; Appendix~\ref{app:convention} reports such cells under both conventions and under the clamp-free estimator.

\textbf{Muon:} Muon ~\citep{jordan2024muon} applies Newton–Schulz orthogonalized momentum to chosen 2-D weight matrices. Where we use it (\S\ref{sec:mech}), it has the same peak learning rate, schedule, and decoupled weight decay as AdamW, and every parameter not on Muon stays on AdamW.

\section{Convention Defect and Dual-Reporting}
\label{app:convention}

\begin{wrapfigure}[18]{r}{0.45\linewidth}
\vspace{-0.6\baselineskip}
\centering
\includegraphics[width=0.85\linewidth]{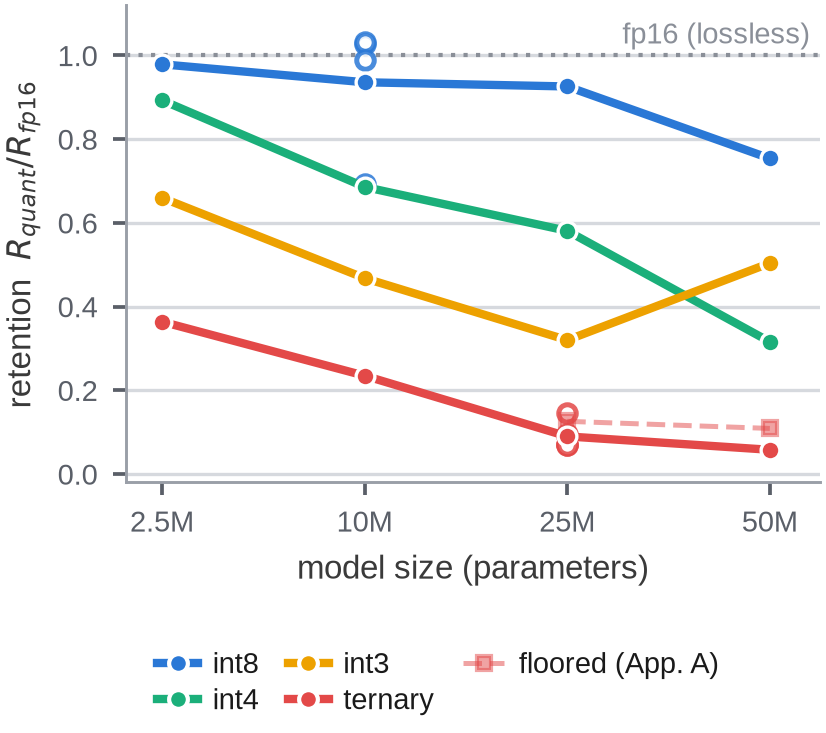}
\caption{Retention at every cell of Table~\ref{tab:grid}, under the standard cosine protocol. Dashed: the ternary cells under the per-attribute floor. Open markers: the seeds of the two cells with more than two.}
\label{fig:retcurve}
\end{wrapfigure}

\textbf{Denominator:} $P$ counts the int4 embedding and output head at full count, so at ``25M'' $P=27,341,568$ rather than $25\times10^6$, with no bit-width weighting anywhere. The fp16 references (Table~\ref{tab:grid}) are consistent with the ${\approx}2$ bits/param of \citet{allenzhu2024capacity} from 10M up; at 2.5M those two matrices are nearly two-fifths of the model, and fp16 is far from saturating its corpus, so 2.5M retentions are ratios of two under-filled models. Excluding them would raise every $R$ by their share of the parameters, from $9.7\%$ at 50M to $39.2\%$ at 2.5M, putting the fp16 references at $1.43/2.04/1.97/1.79$ and accounting for most of the low 2.5M value; retention at matched size is identical under either choice.

Definition~4.3 of \citet{allenzhu2024capacity} sums the value term over the six attributes and imposes no non-negativity: a model that is anti-predictive on an attribute is scored negatively. Taken literally, it therefore gives $R=-0.23$, $-2.33$, and $-2.51$ for the three 50M ternary cosine seeds, which cannot form a retention ratio; some floor is unavoidable, and where to put it is ours, not theirs. Clamping the aggregate once, as we do throughout, zeros genuine partial knowledge stored in the other five attributes as soon as one attribute goes strongly anti-predictive, which happens systematically after the collapse. Flooring per attribute instead keeps that knowledge but corrects six noisy terms rather than one, biasing upward wherever an attribute sits near zero. We report both conventions wherever a collapsed checkpoint is involved: at 25M ternary cosine, retention is $0.090$ under the aggregate clamp and $0.126$ under the per-attribute floor; at 50M, $0.058$ versus $0.109$. Qualitative conclusions are unchanged under either, but several magnitude claims from an earlier version of this analysis (a ``$4\times$ budget is worse than $1\times$'' finding in particular) were artifacts of the aggregate clamp and are retracted. A standing audit flags any cell with a raw negative term before it enters a table.

\textbf{Per-seed 25M ternary endpoints:} Cosine: $0.122/0.155/0.118/0.164/0.253/0.120$ (the two \texttt{grid\_25m} seeds, the two \texttt{tern25} seeds, and the two AdamW seeds of Table~\ref{tab:opt}); WSD 10\%: $0.423/0.421/0.381/0.410/0.426$.

Table~\ref{tab:conv} collects every quantity in the paper whose value depends on this choice, under all three estimators: the aggregate clamp used throughout, the per-attribute floor, and the clamp-free compression estimator of \citet{morris2025memorization} (Appendix~\ref{app:morris}). The conventions differ wherever an attribute ends anti-predictive, so every collapsed cell reads higher under the floor: besides the $\dagger$ cells, the collapsed optimizer-placement arms of Table~\ref{tab:opt} and the recovery arms read 0.03-0.05 higher (0.158 against 0.210 for Muon on the body, seed 0; 0.126 against 0.165 for the embedding and head rate $/10$, seed 1). Pre-collapse cells at 25M and 50M agree within 0.01 under the two conventions, and 2.5M cells agree within 0.035; no ordering changes in any table.

\begin{table}[h]
\caption{Every quantity that depends on the clamp convention, under all three estimators: the $\dagger$ cells of Tables~\ref{tab:grid}, \ref{tab:dwell} and \ref{tab:wsd}, with pre-collapse reference cells for calibration; held-rate rows are the 500-exposure checkpoints. Dashes: not run. $\P$: Morris value over the two grid seeds only. $\ddagger$: one 50M cosine seed scores $-1.78$ under our Morris implementation, which caps held-out cross-entropy at $\log V$ (uncapped, $+0.643$); wherever negative, the column is a signed compression gain (Appendix~\ref{app:morris}).}
\label{tab:conv}
\centering
\small
\begin{tabular}{llccc}
\toprule
cell & quantity & aggregate clamp & per-attribute floor & Morris \\
\midrule
10M ternary cosine & retention & 0.235 & 0.235 & 0.200 \\
25M ternary cosine & retention & 0.090 & 0.126 & 0.185$^{\P}$ \\
50M ternary cosine & retention & 0.058 & 0.109 & $-0.250^{\ddagger}$ \\
25M ternary WSD 10\% & retention & 0.240 & 0.240 & 0.224 \\
25M ternary WSD 33\% & retention & 0.150 & 0.167 & 0.171 \\
50M ternary WSD 33\% & retention & 0.067 & 0.125 & $-0.484^{\ddagger}$ \\
50M ternary WSD 10\% & retention & 0.247 & 0.247 & 0.253 \\
50M int4 cosine & retention & 0.316 & 0.490 & 0.583 \\
25M int3 cosine & retention & 0.320 & 0.320 & 0.423 \\
25M int3 WSD 10\% & retention & 0.520 & 0.520 & 0.462 \\
\midrule
held $2.0\times10^{-4}$ s0 & final $R$ & 0.101 & 0.183 & 0.558 \\
held $2.0\times10^{-4}$ s1 & final $R$ & 0.191 & 0.203 & 0.371 \\
held $1.5\times10^{-4}$ s1 & final $R$ & 0.109 & 0.202 & 0.064 \\
held $1.2\times10^{-4}$ s0 & final $R$ & 0.105 & 0.195 & 0.166 \\
held $1.2\times10^{-4}$ s1 & final $R$ & 0.182 & 0.211 & $-1.646^{\ddagger}$ \\
held $1.0\times10^{-4}$ s0 & final $R$ & 0.110 & 0.201 & 0.438 \\
held $1.0\times10^{-4}$ s1 & final $R$ & 0.158 & 0.215 & 0.601 \\
held $0.8\times10^{-4}$ s0 & final $R$ & 0.111 & 0.207 & $-1.129^{\ddagger}$ \\
held $0.8\times10^{-4}$ s1 & final $R$ & 0.195 & 0.214 & 0.362 \\
held $0.5\times10^{-4}$ s1 & final $R$ & 0.188 & 0.233 & 0.636 \\
held $3.0\times10^{-4}$ s1 & final $R$ & 0.107 & 0.198 & 0.537 \\
\midrule
constant LR s0 (pre-collapse) & final $R$ & 0.338 & 0.338 & 0.557 \\
\bottomrule
\end{tabular}
\end{table}

\section{Depth and Recipe Controls}
\label{app:controls}

\textbf{Light-load cells:} At $N=371$k, ternary reaches $R=0.363/0.347/0.355$ at 25M and $0.319/0.320$ at 50M; the matched-load fp16 references are $0.724/0.713$ and $0.408/0.409$, so the light-load ratios are $0.49$ and $0.78$, compared to full-load ternary values of $0.155$ and $0.094$. \emph{Step-matched light load:} the 25M light load ternary model trained for 3029 exposures under cosine takes the same number of optimizer steps as the full-load run and the same number inside the hazard window. It never meets the collapse criterion and ends at $R=0.392$ and $0.385$ (two seeds, evaluated every 75 exposures), above the 1000-exposure light-load values.

Rebuilding the 10M and 25M ternary cells with matched parameter counts but doubled depth helps at 25M and hurts at 10M, and the recovered 25M point stays below the 10M baseline, which is not the signature of a shape confound. Five further interventions (higher peak LR, the two-stage LR-and-weight-decay recipe of \citet{spectra2024trilm}, a $4\times$ budget, fine-tuning from a trained fp16 checkpoint, and fp16 in the embedding and head) all reach the same $0.35$--$0.40$ level; the $4\times$ budget runs then collapse late under cosine exactly as the standard runs do ($R=0.098/0.105$), while the fine-tuning runs do not collapse at all and finish at $R=0.402/0.371$. The fp16 embedding and head variant has since been run at full load, the condition that collapses: with the embedding and head at fp16 instead of int4 and everything else unchanged, one of two cosine seeds collapsed near exposure 625 and partially recovered ($R=0.278$) while the other never collapsed ($R=0.325$), against six of six int4 seeds at the post-collapse level. The lattice is therefore not necessary for the failure; whether higher precision there lowers the rate is suggested but not established at $n{=}2$. The fine-tuning run is instructive on its own: initializing from an fp16 checkpoint storing $R=1.68$ destroys roughly 90\% of the inherited knowledge within 25 exposures and converges to the from-scratch plateau, whereas the int4-adapted warm start of the curriculum in \S\ref{sec:fix} builds past it: lattice-adjacent initializations transfer, distant ones do not.

\begin{figure}[h]
\centering
\includegraphics[width=\linewidth]{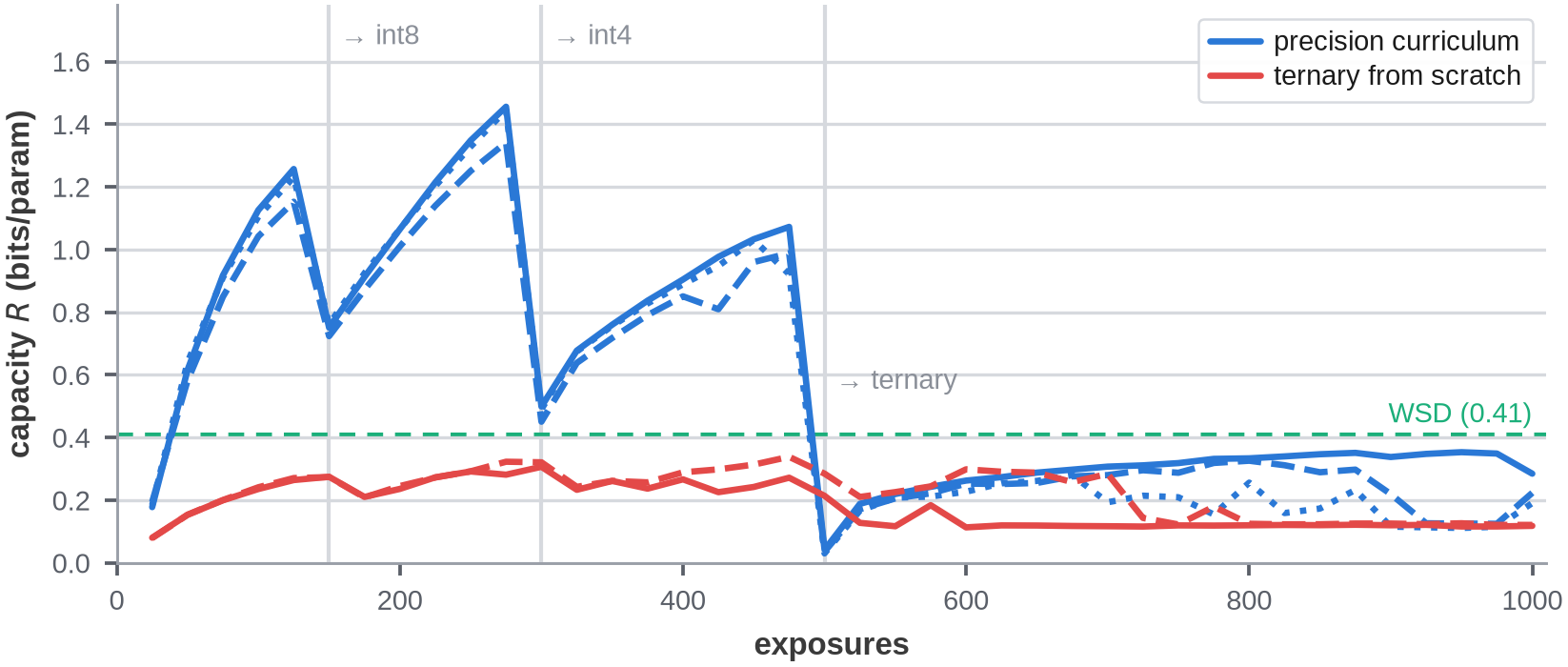}
\caption{Precision curriculum (three seeds) against ternary-from-scratch on the identical learning-rate path (two seeds); stage boundaries marked, the WSD level dashed for reference.}
\label{fig:curriculum}
\end{figure}

\section{Gradient-Estimator Controls}
\label{app:ste}

Every run elsewhere in the paper uses the clipped straight-through estimator. A reviewer of an earlier draft asked whether the collapse is an artifact of that choice. It is not, but the dependence runs in one direction only (Table~\ref{tab:ste}).

\textbf{From scratch, the alternatives do not collapse:} Neither the identity estimator nor stochastic rounding meets the criterion under cosine, on two seeds each. Both still gain from a cooldown, so the schedule ordering survives the estimator change. Slower learning does not explain the survival: stochastic rounding reaches $R=0.33$, the level from which every clipped cosine seed falls, and holds it to the end. The identity estimator peaks lower, at $0.27$, so for that arm we cannot separate the estimator from the state it never reaches.

\textbf{Once a clipped-trained model exists, the estimator stops mattering:} We branch the pre-collapse clipped parents to a held rate of $1\times10^{-4}$ and run the hold with the identity backward pass. Both seeds collapse at onsets 275 and 245, within the noise of their clipped siblings. Identity-trained parents held the same way do not collapse and end above their parents. We therefore read the clipped estimator as building the state that collapses rather than driving the collapse itself, on two lineages.

\begin{table}[h]
\caption{The collapse needs a clipped-trained model, not a clipped backward pass. Final $R$ at 25M ternary, full load; the upper block trains from scratch for 1000 exposures, the lower holds a parent at $1\times10^{-4}$ for 300 exposures with the stated backward pass.}
\label{tab:ste}
\centering
\small
\begin{tabular}{llcc}
\toprule
 & estimator & final $R$ & collapse \\
\midrule
\multirow{3}{*}{cosine, from scratch} & clipped (six seeds) & 0.118--0.253 & 6 of 6 \\
 & identity & 0.245 / 0.264 & no \\
 & stochastic rounding & 0.327 / 0.322 & no \\
\midrule
\multirow{3}{*}{WSD 10\%, from scratch} & clipped & 0.410 / 0.426 & no \\
 & identity & 0.308 / 0.324 & no \\
 & stochastic rounding & 0.343 & no \\
\midrule
\multirow{3}{*}{held at $1\times10^{-4}$} & clipped parent, clipped hold & 0.117 / 0.126 & both \\
 & clipped parent, identity hold & 0.113 / 0.109 & both \\
 & identity parent, identity hold & 0.290 / 0.307 & no \\
\bottomrule
\end{tabular}
\end{table}

\section{Onset Learning Rates and the Temperature Control}
\label{app:onset}

\begin{figure}[h]
\centering
\includegraphics[width=\linewidth]{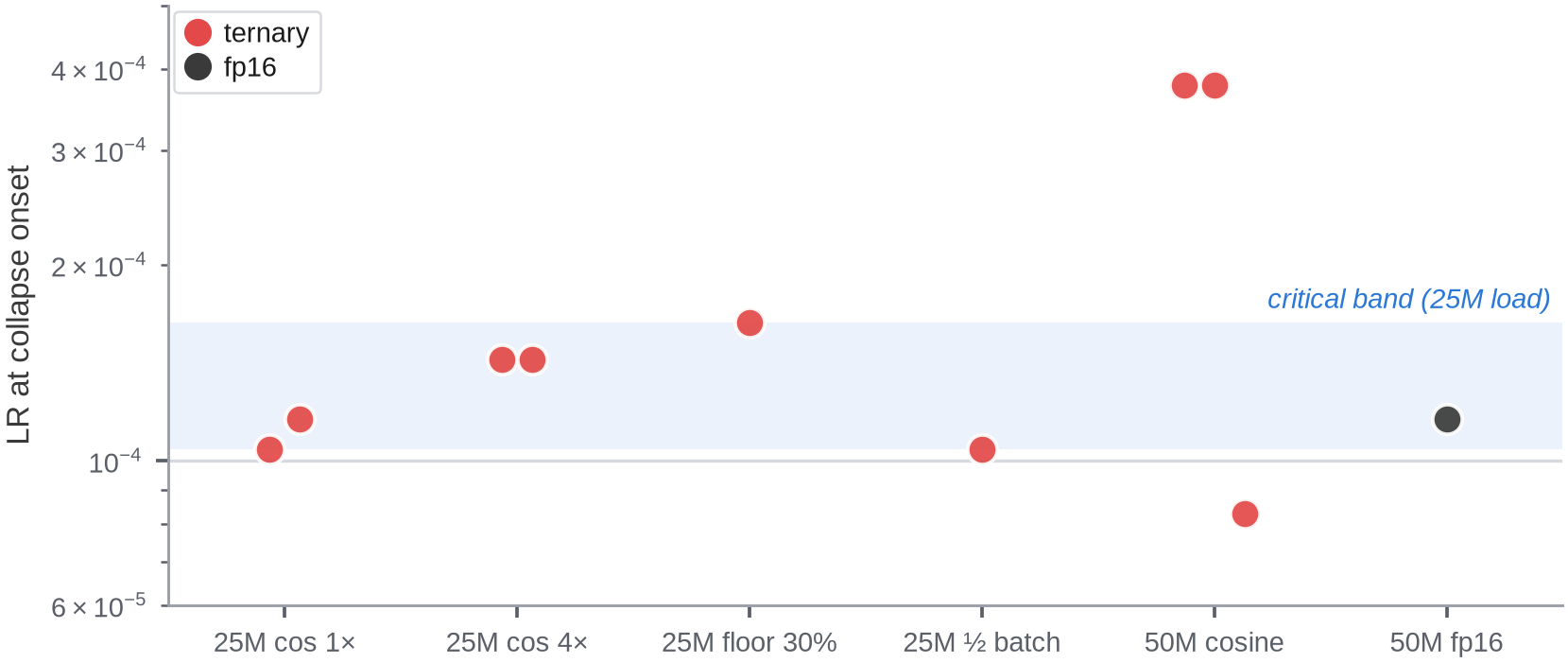}
\caption{Learning rate at collapse onset for every collapsing run with a logged trace. The shaded band spans the 25M values, which \S\ref{sec:dwell} links to the hazard at each rate and how long cosine spends there.}
\label{fig:onset}
\end{figure}

Reading off the learning rate at the first evaluation where capacity drops below half its running peak, five 25M crashing runs span a $4\times$ change in training length and a $3\times$ change in floor, crossing into collapse between $1.0$ and $1.6\times10^{-4}$; two of three 50M seeds cross at $3.8\times10^{-4}$ at 34\% of the schedule, and the crashed 50M fp16 seed at 70\%, inside the 25M band (Figure~\ref{fig:onset}). A gradient-noise account predicts a band in learning rate divided by batch size; halving the batch (96 instead of 192) leaves onset at the same $1.04\times10^{-4}$ rather than the $0.5$--$0.8\times10^{-4}$ predicted, a single replicate that disfavors the temperature account without ruling it out.

\section{Dwell Map and Recovery: Traces}
\label{app:dwell}

\begin{figure}[h]
\centering
\includegraphics[width=\linewidth]{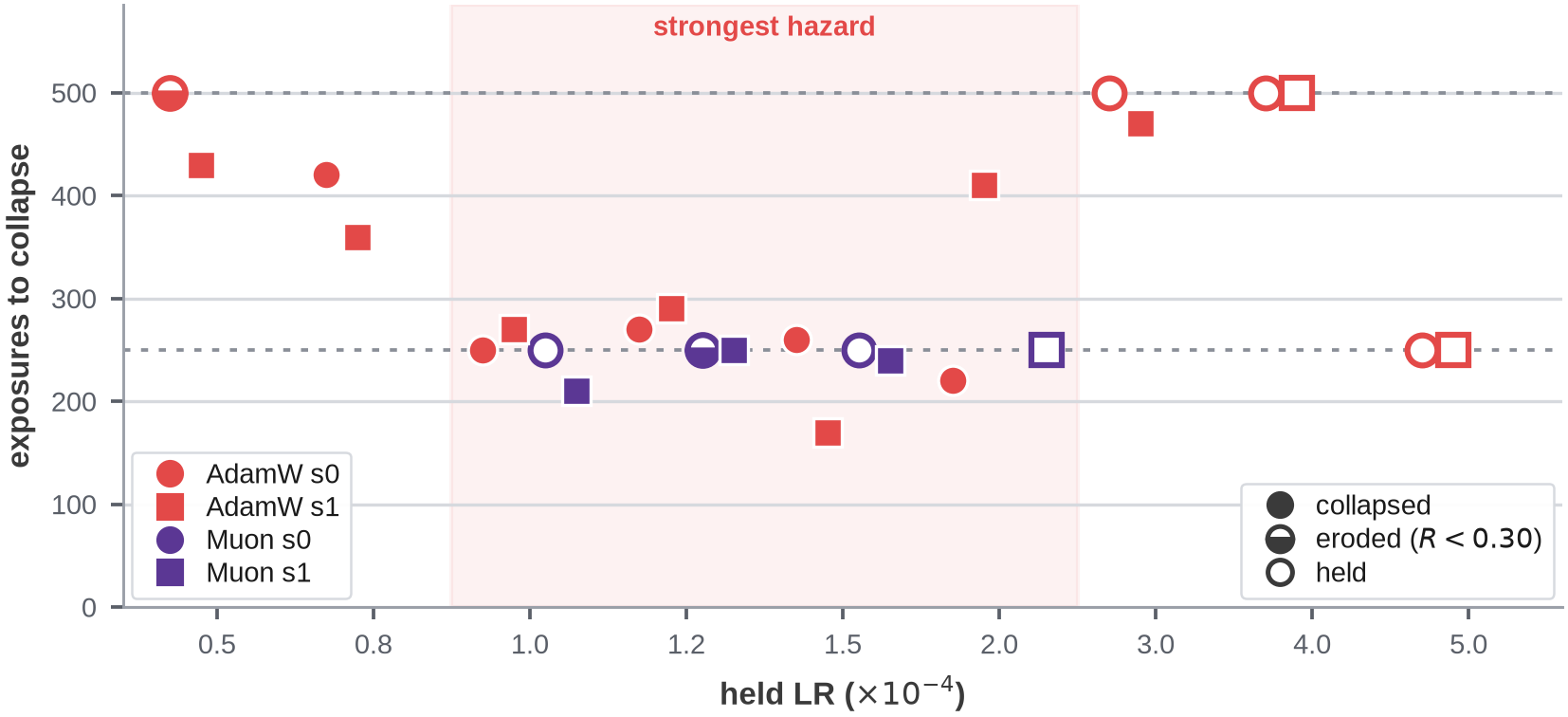}
\caption{Hazard map from Table~\ref{tab:dwell}: dwell exposures to sustained collapse at a held learning rate. Color: optimizer; shape: seed; filled: collapse; half-filled: survived but eroded; open: held capacity (censored at the hold length, dotted). Shading marks where the hazard is strongest.}
\label{fig:hazard}
\end{figure}

\begin{figure}[h]
\centering
\includegraphics[width=\linewidth]{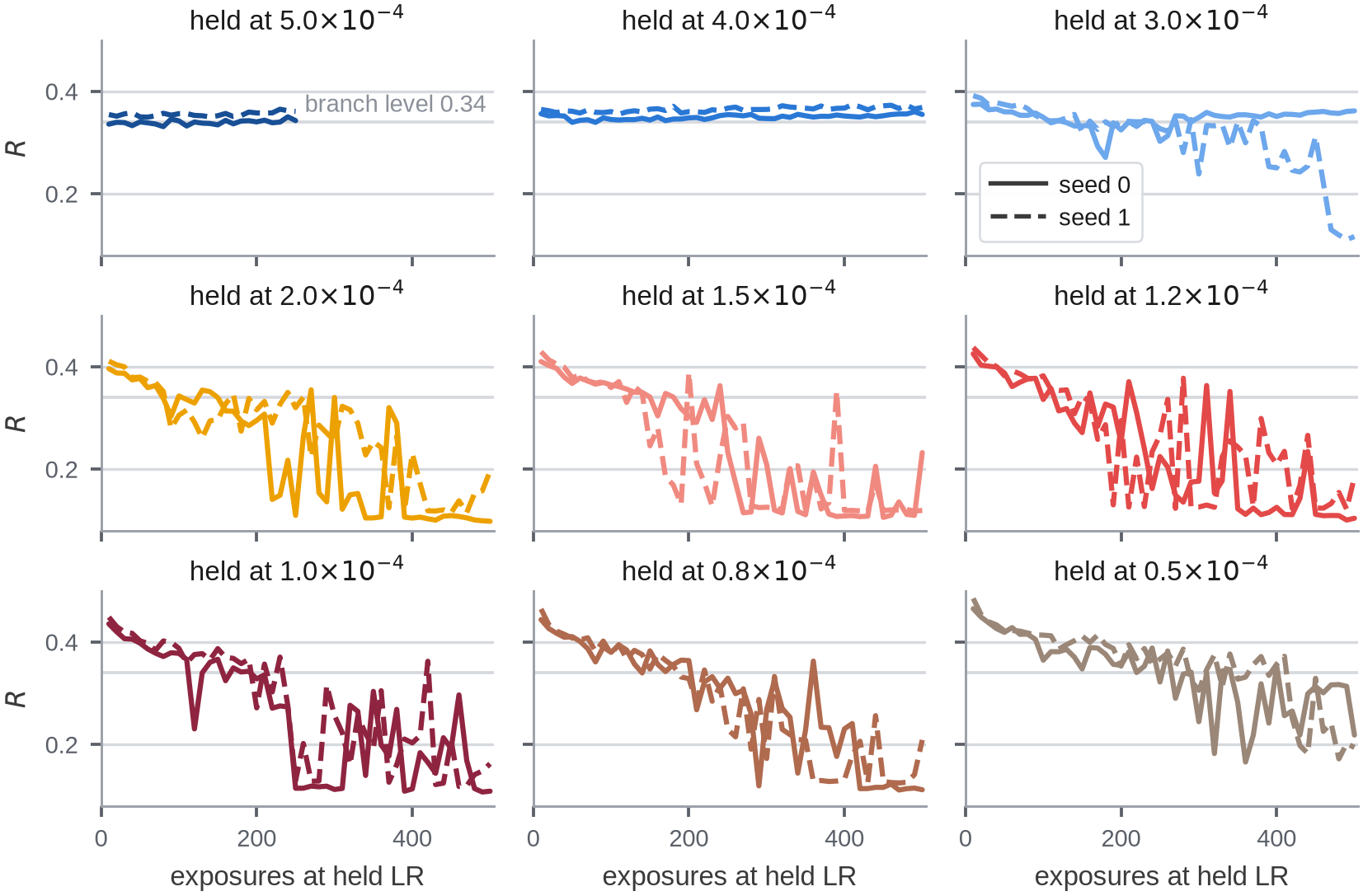}
\caption{Held-LR runs of Table~\ref{tab:dwell}, branched from the pre-collapse constant-LR checkpoints, one panel per held rate.}
\label{fig:dwellfan}
\end{figure}

\begin{figure}[h]
\centering
\includegraphics[width=\linewidth]{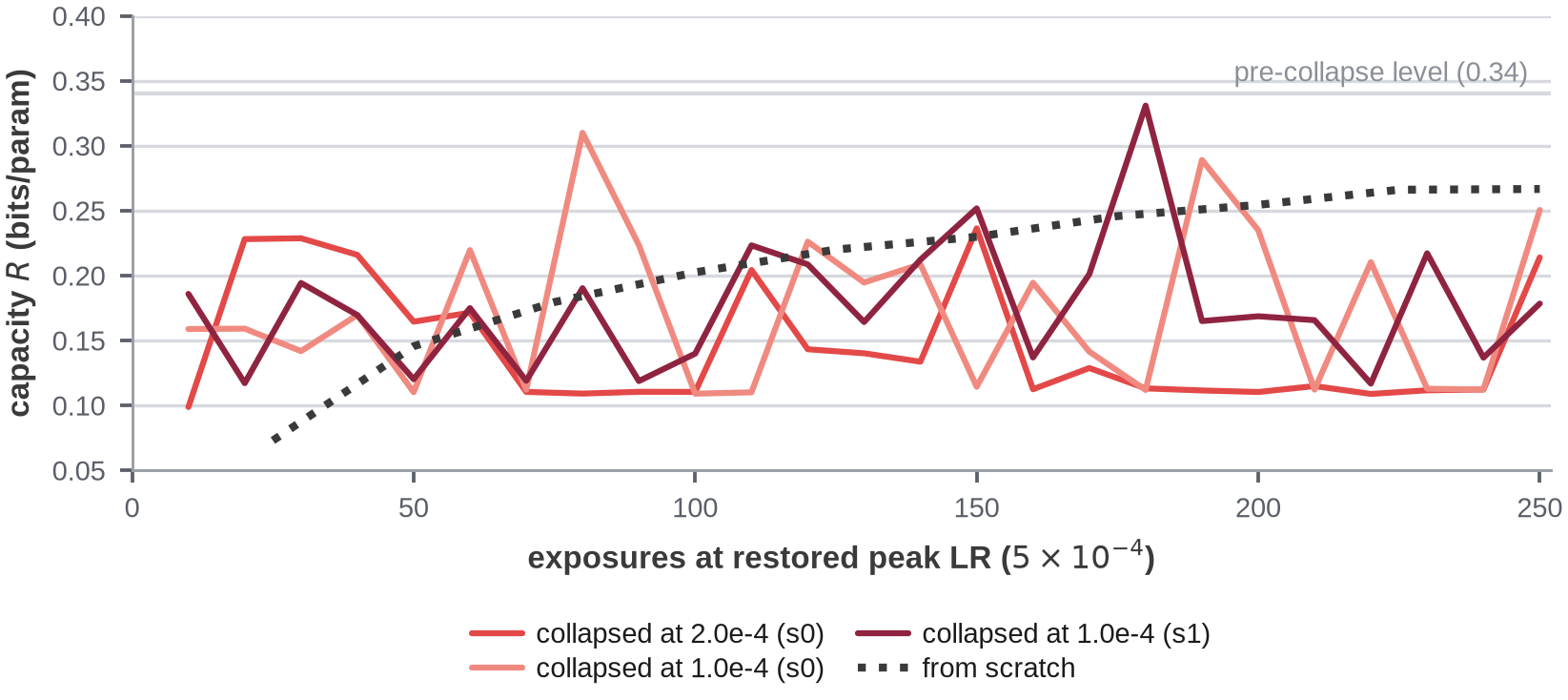}
\caption{Recovery: three collapsed checkpoints returned to the safe constant rate $5\times10^{-4}$ for 250 exposures, against a fresh model at the same rate.}
\label{fig:rescue}
\end{figure}

\textbf{Optimizer-state reset:} Branching the same three collapsed checkpoints back to $5\times10^{-4}$ with the AdamW moments and step counters dropped for all parameters, for the embedding and head only, or for the body only, gives the same outcome as keeping them: every arm spends its last 50 exposures flipping between the collapsed level and about $0.3$, and the spread across arms equals the spread between two replicates of the identical unreset configuration. Stale optimizer state does not hinder recovery.

\begin{figure}[h]
\centering
\includegraphics[width=\linewidth]{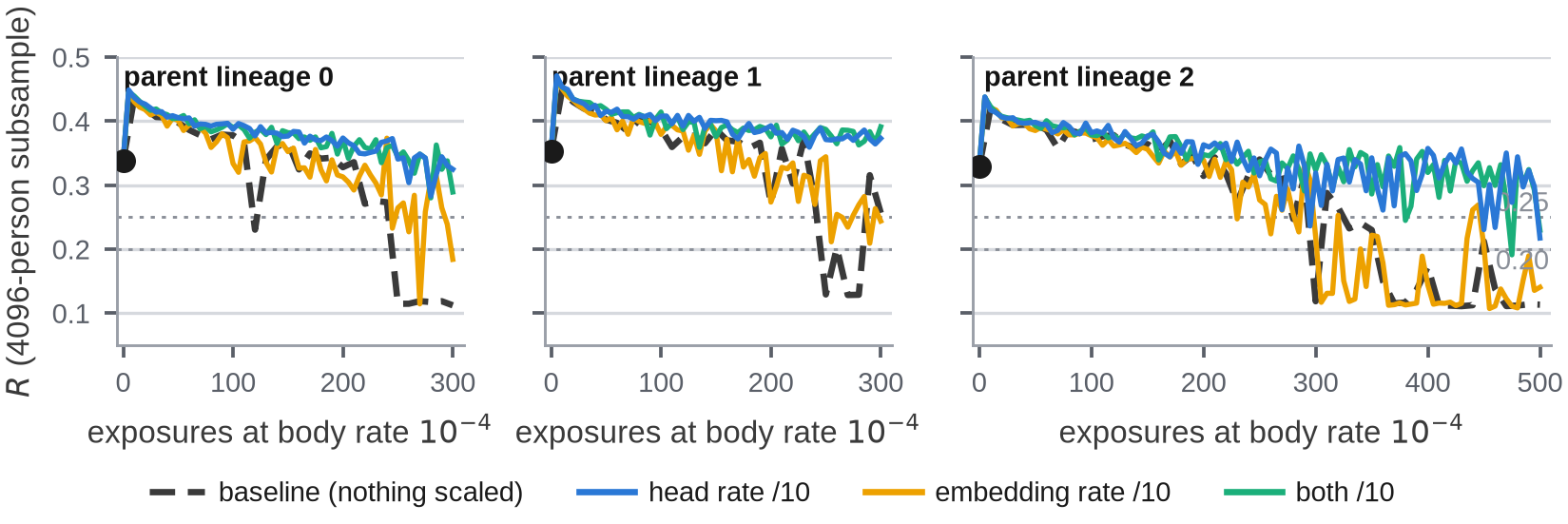}
\caption{Interface split on three parent lineages. Each panel branches one pre-collapse 25M ternary parent (filled marker) to a held body rate of $10^{-4}$, evaluated every five exposures on the subsample; the dashed baseline scales nothing, the coloured arms divide the head, the embedding, or both by ten with weight decay compensated. Dotted lines mark the 0.25 and 0.20 levels of the collapse criterion.}
\label{fig:interface}
\end{figure}

\textbf{Interface split on pre-collapse parents:} We call the int4 embedding and output head the model's \emph{interface}. Three AdamW parameter groups (body, embedding, head) receive independent learning-rate multipliers; ``compensated'' raises the group's weight decay by the inverse factor so that $\eta\lambda$ matches the body's. Arms branch from the pre-collapse constant-LR parents (\texttt{opt25\_adamw\_constant\_s0/s1}, $R$ 0.338 and 0.353; \texttt{constlr\_tern\_s2}, 0.329) to a body rate of $1\times10^{-4}$ and evaluate the subsample every five exposures; the baselines are the $1\times10^{-4}$ held-rate branches. We used lineages 0 and 1 (300 exposures) to choose the arms. Before launching lineage 2 (500 exposures, chosen because its baseline collapsed later) the prediction was recorded: \emph{head/10 and both/10 do not meet the collapse criterion within 500 exposures and end with tail-mean $R>0.28$ and tail-mean birth-date loss $<9.6$ nats; embed/10 crosses 0.25 and ends with birth-date loss $>10$.} Three clauses held; the birth-date clause did not (tail 9.88 and 9.88 nats, final evaluations 11.15 and 10.93). Table~\ref{tab:a3} and Figure~\ref{fig:interface} give every arm.

\begin{table}[h]
\centering\small
\caption{Interface-split arms: full-corpus final $R$, peak and minimum of the subsample trace, the exposure at which the trace first falls below 0.25 and the criterion onset (``--'' if never within the horizon), and tail means over the last 50 exposures. Horizons are 300 exposures for lineages 0 and 1 and 500 for lineage 2, whose baseline concatenates the 250-exposure hold and its extension.}
\label{tab:a3}
\setlength{\tabcolsep}{4pt}
\begin{tabular}{llccccccc}
\toprule
Lineage & Arm & Final $R$ & Peak & Min & First $<$0.25 & Onset & Tail $R$ & Tail BD nll \\
\midrule
0 & baseline & 0.117 & 0.435 & 0.115 & 120 & -- & 0.254 & 10.56 \\
0 & head /10 & 0.322 & 0.449 & 0.283 & -- & -- & 0.328 & 9.32 \\
0 & embed /10 & 0.179 & 0.447 & 0.114 & 245 & -- & 0.247 & 10.47 \\
0 & both /10 & 0.289 & 0.446 & 0.280 & -- & -- & 0.330 & 9.30 \\
1 & baseline & 0.126 & 0.449 & 0.129 & 250 & -- & 0.287 & 10.24 \\
1 & head /10 & 0.371 & 0.471 & 0.359 & -- & -- & 0.374 & 8.86 \\
1 & embed /10 & 0.244 & 0.465 & 0.209 & 255 & -- & 0.247 & 10.85 \\
1 & both /10 & 0.390 & 0.468 & 0.359 & -- & -- & 0.377 & 8.79 \\
2 & baseline & 0.113 & 0.419 & 0.111 & 280 & 360 & 0.125 & 14.06 \\
2 & head /10 & 0.213 & 0.438 & 0.217 & 295$^*$ & -- & 0.294 & 9.88 \\
2 & embed /10 & 0.139 & 0.437 & 0.107 & 230 & 305 & 0.131 & 12.45 \\
2 & both /10 & 0.226 & 0.438 & 0.191 & 380$^*$ & -- & 0.291 & 9.88 \\
\bottomrule
\end{tabular}
\\[2pt]{\footnotesize $^*$Single-evaluation dip that recovered. The lineage-0 and lineage-1 baselines and embedding arms flicker on the evaluation grid and post no two consecutive values below 0.20 within their horizons, although their full-corpus final values are at the collapsed level.}
\end{table}

\textbf{Collapse and birth date:} Across the thirteen held-rate arms that fall below $0.2$ after exceeding $0.25$, final capacity and final birth date loss rank together at $\rho=-0.96$. These arms end at $10.4$--$22.7$ nats on birth date, against $8.6$--$8.7$ for the arms that never collapse. Because $R$ contains the birth-date term, this describes the collapse rather than evidencing its mechanism.

\section{Per-Token Audit Protocol}
\label{app:audit}

Each archived checkpoint is scored on a corpus generated with that checkpoint's own seed, since the biography population depends on the seed, and scoring a model against another seed's population places every attribute at chance. We evaluate under an fp32 forward with two batch sizes as a numerical control, and additionally under bf16 to expose logits too large to resolve. The audit records, per attribute and per token position within the birth date, the negative log-likelihood, the top-1 accuracy of the final token, and $\max|z|$ over the position's vocabulary entries. It also records the mean $\ell_2$ norm of the output-head rows for the 12-month tokens, the 28-day tokens, the 200-year tokens, and all 2985 rows. The uniform baseline is $\ln 12 = 2.485$ nats for the month, $\ln 28 = 3.332$ for the day and $\ln 200 = 5.298$ for the year; a model scoring above its baseline at a position is anti-predictive there, which is what the aggregate clamp responds to (Appendix~\ref{app:convention}).

\begin{table}[h]
\caption{A collapsed checkpoint has lost nothing except the month. Per-attribute loss in nats under an fp32 forward, on the corpus of each checkpoint's own seed; $\max|z|$ is over the four date-token positions and ``month row'' is the mean norm of the 12 month rows of the output head. Pre-collapse rows are the constant-LR parents and the two settings that do not collapse; collapsed rows meet the \S\ref{sec:setup} criterion.}
\label{tab:parity}
\centering
\footnotesize
\setlength{\tabcolsep}{4pt}
\begin{tabular}{lccccccccc}
\toprule
checkpoint & $R$ & month & day & year & year top-1 & name & city & $\max|z|$ & month row \\
\midrule
constant LR s0 & 0.338 & 2.58 & 3.20 & 2.97 & 0.322 & 16.81 & 4.59 & 356 & 18.6 \\
constant LR s1 & 0.353 & 2.59 & 3.13 & 3.06 & 0.305 & 16.70 & 4.45 & 379 & -- \\
WSD 10\% s0 & 0.410 & 2.70 & 3.04 & 2.48 & 0.395 & 16.55 & 4.42 & 685 & 19.0 \\
head rate $/10$ & 0.371 & 2.57 & 2.73 & 2.86 & 0.325 & 16.76 & 4.63 & 434 & 2.5 \\
\midrule
cosine s1 & 0.120 & 7.22 & 3.18 & 3.00 & 0.309 & 16.89 & 4.48 & 10{,}288 & -- \\
cosine (grid s0) & 0.122 & 9.07 & 3.32 & 2.61 & 0.378 & 16.87 & 4.42 & 9{,}752 & 39.7 \\
cosine (grid s1) & 0.155 & 6.04 & 3.14 & 2.97 & 0.326 & 16.79 & 4.39 & 11{,}230 & 42.3 \\
held $1.0\times10^{-4}$ s0 & 0.117 & 7.21 & 3.17 & 2.72 & 0.362 & 16.96 & 4.39 & 5{,}781 & -- \\
held $1.0\times10^{-4}$ s1 & 0.126 & 7.22 & 3.09 & 2.81 & 0.344 & 16.71 & 4.29 & 5{,}783 & -- \\
held $1.2\times10^{-4}$ s0 & 0.105 & 10.61 & 3.18 & 2.80 & 0.337 & 17.14 & 4.43 & 12{,}710 & -- \\
held $2.0\times10^{-4}$ s0 & 0.101 & 15.76 & 3.19 & 2.89 & 0.325 & 17.23 & 4.49 & 16{,}283 & 53.9 \\
held $3.0\times10^{-4}$ s1 & 0.107 & 10.82 & 3.12 & 3.03 & 0.310 & 16.93 & 4.43 & 10{,}007 & -- \\
embedding rate $/10$ & 0.126 & 16.32 & 3.29 & 2.98 & 0.318 & 16.61 & 4.57 & 12{,}079 & -- \\
\bottomrule
\end{tabular}
\end{table}

\textbf{A census across precisions and sizes:} Table~\ref{tab:census} scores one final checkpoint per cell. The monthly loss sits at or above its uniform value in every cell but two, and the rows that predict it grow as precision falls while the average row does not. A 10\% cooldown holds every cell at the uniform monthly loss and keeps the largest date logit in the hundreds. The day and the year of the same date are learned almost perfectly in fp16, at $0.48$ and $0.29$ nats. The two exceptions support the account rather than weakening it. Of the two int8 cosine seeds, the one that learns the month carries a month row norm of $8.0$, and the one that does not carries $36.8$; the 50M int3 seed that partly learns it carries the smallest date logits of any 50M cell below int8, $1{,}064$ against $12{,}587$ and $14{,}376$.

Two readings sit outside the table. At 10M, where the corpus holds 371k people rather than 1.12M, fp16 does learn the month, reaching top-1 $0.639$ at $1.48$ nats. And at 50M, the same growth appears in the person's name, the other position whose loss cannot be driven to zero. The crashed fp16 seed and the int8 seed both store negative bits of the name, at $-25$ and $-15$. Their largest name logits reach $9{,}423$ and $4{,}237$, against $833$ for the fp16 seed that does not crash. No 25M checkpoint shows it at any precision.

\begin{table}[t]
\caption{The month loss sits at or above its uniform value of $\ln 12 = 2.485$ nats in all but two checkpoints, and the output-head rows that predict it grow as precision falls. One final checkpoint per row under an fp32 forward, scored on its own seed's corpus. ``month row'' and ``all rows'' are mean $\ell_2$ norms; $\max|z|$ is over the four date positions.}
\label{tab:census}
\centering
\footnotesize
\setlength{\tabcolsep}{5pt}
\begin{tabular}{llcccccr}
\toprule
size & precision & $R$ & month & month top-1 & month row & all rows & $\max|z|$ \\
\midrule
\multicolumn{8}{l}{\emph{cosine}} \\
25M & fp16 & 1.722 & 2.70 & 0.071 & 11.2 & 2.70 & 1{,}878 \\
25M & int8 & 1.670 & 2.16 & 0.341 & 8.0 & 2.05 & 1{,}132 \\
25M & int8 & 1.515 & 3.42 & 0.067 & 36.8 & 3.62 & 6{,}705 \\
25M & int4 & 1.016 & 4.67 & 0.068 & 33.9 & 1.84 & 4{,}145 \\
25M & int4 & 0.982 & 5.84 & 0.067 & 35.6 & 1.73 & 5{,}109 \\
25M & int3 & 0.580 & 7.19 & 0.068 & 37.9 & 1.82 & 4{,}242 \\
25M & int3 & 0.522 & 6.82 & 0.077 & 44.6 & 1.84 & 7{,}257 \\
25M & ternary & 0.122 & 9.07 & 0.066 & 39.7 & 2.17 & 9{,}752 \\
25M & ternary & 0.155 & 6.05 & 0.077 & 42.3 & 2.29 & 11{,}230 \\
50M & fp16 & 1.609 & 2.92 & 0.080 & 22.4 & 5.72 & 6{,}414 \\
50M & fp16 (crashed) & 1.374 & 2.94 & 0.076 & 23.8 & 9.77 & 11{,}233 \\
50M & int8 & 1.220 & 5.11 & 0.080 & 33.6 & 6.56 & 9{,}060 \\
50M & int4 & 0.512 & 13.82 & 0.070 & 45.9 & 2.18 & 12{,}587 \\
50M & int3 & 0.815 & 2.38 & 0.163 & 19.9 & 2.52 & 1{,}064 \\
50M & ternary & 0.108 & 11.94 & 0.074 & 40.3 & 3.07 & 14{,}376 \\
\midrule
\multicolumn{8}{l}{\emph{WSD, 10\% cooldown}} \\
25M & int8 & 1.702 & 2.56 & 0.113 & 4.9 & 2.63 & 284 \\
25M & int4 & 1.215 & 2.61 & 0.077 & 12.2 & 2.54 & 487 \\
25M & int3 & 0.904 & 2.63 & 0.074 & 13.6 & 2.74 & 449 \\
25M & ternary & 0.423 & 2.59 & 0.075 & 21.4 & 2.92 & 853 \\
50M & ternary & 0.382 & 2.78 & 0.074 & 38.1 & 3.01 & 3{,}049 \\
\bottomrule
\end{tabular}
\end{table}

\textbf{The two forward precisions disagree only where the logits are large:} An fp32 and a bf16 forward over the same weights agree to three decimals on every attribute of a pre-collapse checkpoint. On a collapsed one, they disagree only at the diverged position. The injected tag of Appendix~\ref{app:corpus} reads $20.33$ nats under fp32 and $2.52$ under bf16, within 0.04 nats of the uniform value. Its sibling seed reads $13.09$ against $4.54$, and the two date-last seeds read $22.73$ against $17.65$ and $16.14$ against $10.85$. A stored fact would not move with the accumulator width, so we read the gap as logits bf16 cannot resolve rather than as damage.

\textbf{The flicker is the quantized head changing state:} Two evaluations break the otherwise monotone rise of the month logits through a cosine run. At exposure 775, the largest month logit falls from $6{,}937$ to $27$. At 925, it falls from $10{,}210$ to $365$. The month loss returns to its uniform value at both, and the month row norms do not change across either pair of snapshots. The latent weights, therefore, did not move, and the int4 head rounded to a different state. The two-evaluation collapse criterion in \S\ref{sec:setup} averages over this.

\section{Corpus Variants}
\label{app:corpus}

Each variant resizes $N$ so the corpus holds 2.25 bits per parameter (Table~\ref{tab:variantn}). Each adds vocabulary entries only while it is active, so the default vocabulary and $P$ remain unchanged.

\begin{description}[leftmargin=1em,labelsep=0.5em,itemsep=2pt,topsep=4pt,parsep=0pt,font=\normalfont\itshape]
\item[No birth date] drops the attribute and stops scoring it. Three seeds reach $R=0.372$, $0.355$ and $0.372$.
\item[Keep date] drops university and major instead, resizing $N$ by about the same amount, and serves as the control for that resizing.
\item[No month] renders the date without its month. Here $\max|z|$ on the date tokens stays between 78 and 202.
\item[Month from name] sets the month to the index of the first name modulo 12. The month becomes a deterministic function of text the model already predicts.
\item[Date last] moves the date sentence to the end of the biography and leaves its content unchanged. The birth city, now the first value in the text, is predicted at top-1 $0.98$, while the divergence stays on the date.
\item[Month 200-way] replaces the 12 month names with 200 tokens drawn uniformly. The model partly learns the wider month, reaching top-1 $0.37$ on the last date token.
\item[Inject tag] drops the birth date and appends one sentence carrying a 12-way tag, a hash of the person's name. The tag is deterministic given the name and carries no structure the model learns at this scale. It draws from its own generator, so it never consumes the shared population random state.
\item[Key--value corpus] replaces biographies with six symbol-valued fields drawn from 256 options each, under four sentence templates.
\end{description}

\textbf{A cooldown lifts every variant it was run on:} Four variants also ran a warmup-stable-decay schedule with a 10\% cooldown, and each one ends above its cosine twin (Table~\ref{tab:variantn}). The lift does not depend on whether the variant collapses under cosine.

\textbf{The key--value corpus has no usable fp16 reference:} Its fp16 cosine run rises to $R=0.241$ by exposure 150 and then erodes to $0.109$, with code loss climbing from $30.5$ to $32.5$ nats. The cooldown arm has the same shape, peaking at $0.507$ and ending at $0.270$. We do not know why full precision fails here, so retention is undefined on this corpus, and we compare ternary schedules within it instead.

\begin{table}[h]
\caption{The re-sizing each variant required, and its cooldown arm where one was run. $N$ is chosen so that every corpus holds 2.25 bits per parameter; cosine values are in Table~\ref{tab:corpus}. Dashes: not run.}
\label{tab:variantn}
\centering
\small
\begin{tabular}{lrl}
\toprule
variant & $N$ & WSD 10\% final $R$ \\
\midrule
unmodified & 1{,}123{,}724 & 0.410 / 0.426 \\
keep date & 1{,}561{,}465 & -- \\
no birth date & 1{,}610{,}874 & 0.412 / 0.408 \\
no month & 1{,}204{,}834 & 0.411 \\
month from name & 1{,}204{,}834 & 0.424 \\
date last & 1{,}123{,}724 & -- \\
month 200-way & 1{,}053{,}303 & -- \\
inject tag & 1{,}468{,}756 & -- \\
key--value corpus & 1{,}053{,}325 & 0.403 \\
\bottomrule
\end{tabular}
\end{table}

\section{Telemetry at the Collapse}
\label{app:telemetry}

\textbf{Body and interface co-adapt during training:} A checkpoint-swap assay, without training, pairs one checkpoint's transformer blocks with another's embedding and output head. Branches held at high rates remain interchangeable with their parent; a branch held near the hazard window loses compatibility while retaining capacity; and a collapsed branch retains none (Table~\ref{tab:swapctl}). Compatibility therefore fails before capacity does, so it is not sufficient for collapse. A capacity-landscape probe (Appendix~\ref{app:landscape}) is consistent with a needle-like ternary basin but is only correlational. A tenfold cut of their learning rate alone does not recover; it differs from the recovered tenfold cell only in their weight decay, since AdamW decays as $w \leftarrow w(1-\eta\lambda)$, so cutting $\eta$ alone also cuts the decay. Freezing them outright is worse than any rate cut, so these layers must keep adapting rather than merely moving less.

\begin{table}[h]
\caption{Optimizer placement $\times$ schedule, 25M ternary full load, two seeds per cell (final $R$). The embedding and output head are int4 in every row, and everything not on Muon ~\citep{jordan2024muon} is on AdamW (Appendix~\ref{app:details}). The rate rows scale the learning rate of the named matrices only; ``$\eta\lambda$ held'' raises their weight decay by the same factor.}
\label{tab:opt}
\centering
\footnotesize
\setlength{\tabcolsep}{4pt}
\begin{tabular}{llll}
\toprule
schedule & placement & final $R$ (s0 / s1) & behavior \\
\midrule
cosine & AdamW everywhere & 0.253 / 0.120 & peak 0.34 at 550--600, collapse \\
cosine & Muon body, AdamW embed/head & 0.158 / 0.204 & peak 0.33 at 625--700, collapse \\
cosine & AdamW, embed/head frozen from 50\% & 0.182 / 0.276 & gradual decline, 600--900 \\
cosine & AdamW, embed/head rate $/10$ & 0.265 / 0.126 & peak 0.31 at 650--675, collapse \\
cosine & AdamW, head rate $/10$ & 0.371 / 0.375 & monotone rise, no collapse \\
cosine & AdamW, embedding rate $/10$ & 0.217 / 0.126 & collapse, onset 825 / 700 \\
cosine & AdamW, embed/head rate $/20$ & 0.323 / 0.334 & monotone rise, no collapse \\
cosine & AdamW, embed/head rate $/10$, $\eta\lambda$ held & 0.335 / 0.333 & monotone rise, no collapse \\
cosine & AdamW body, Muon \emph{only} on embed/head & \textbf{0.350 / 0.347} & monotone rise, no collapse \\
cosine & Muon everywhere & 0.334 / 0.331 & monotone rise, no collapse \\
\midrule
constant & AdamW & 0.338 / 0.353 & stable \\
constant & Muon body & 0.358 / 0.359 & stable \\
WSD 10\% & AdamW & 0.410 / 0.426 & stable, cooldown lift \\
WSD 10\% & Muon body & 0.378 / 0.406 & stable, cooldown lift \\
\bottomrule
\end{tabular}
\end{table}

Ternary-state fractions and flip rate, logged every 3000 steps across every crashing run, do not move at onset: the flip rate drifts smoothly through the window with no spike, and the zero-fraction stays within the noise of its long-run average. The one signature is in the cure: during a WSD cooldown, flip rate falls from about $0.13$ to $0.10$ as capacity rises. The update-to-lattice-step ratio (RMS raw per-step update over local quantization step, median over groups, logged every 1000 steps) has a per-run median between $2\times10^{-4}$ and $1.4\times10^{-3}$ across all 60 instrumented runs, spanning every optimizer, schedule, and run state. At each run's criterion onset, it reads $5.0\times10^{-4}$ (AdamW, mean of two seeds) versus $1.6\times10^{-4}$ (Muon on the body), a $3.1\times$ gap where the onset learning rates differ by $1.3\times$; a WSD cooldown ends at $2.1\times10^{-4}$ (AdamW) and $1.1\times10^{-4}$ (Muon), below both onset values, with no collapse. The flip telemetry covers the ternary body only; the update-ratio telemetry pools the int4 embedding and output head with the body. Neither is layer-resolved, which is the most direct follow-up we did not run.

\textbf{Checkpoint swaps:} With no training, we load a body (all ternary transformer blocks) from one checkpoint and an int4 embedding and output head from another, and run the full-corpus evaluation. Unswapped checkpoints reproduce their recorded $R$ to three decimals, except for the post-recovery checkpoint, which re-evaluates at $0.225$ against a recorded $0.217$; two evaluations of one fixed checkpoint should agree, and we flag the discrepancy rather than explain it. Every swap scores $R=0.000$: pre-collapse body with a collapsed embedding and head and the reverse, for both same-lineage held-LR runs ($2.0$ and $1.0\times10^{-4}$), the post-recovery checkpoint, and the cross-lineage checkpoints that ran Muon on the embedding and head or froze them. Name-prediction loss rises from $16.8$ to $21$--$412$ nats and attribute loss from $29$ to $37$--$531$ nats, with same-lineage swaps at the low end of both: every mixed model is worse than a uniform prior. The pre-collapse same-lineage control that this reading requires is Table~\ref{tab:swapctl}.

\begin{table}[h]
\caption{Pre-collapse same-lineage swap control (evaluation only). Each branch starts from the same parent ($R=0.338$) and is held at a fixed rate for 250 exposures; ``swap'' pairs the parent's body with the branch's embedding and head, and vice versa. Unswapped checkpoints reproduce their recorded $R$.}

\label{tab:swapctl}
\centering
\small
\begin{tabular}{llccc}
\toprule
held LR & dwell-map outcome & own $R$ & swap $R$ (both directions) & \% of parent \\
\midrule
$5.0\times10^{-4}$ & flat & 0.348 & 0.281 / 0.303 & 86 \\
$4.0\times10^{-4}$ & stable & 0.353 & 0.286 / 0.297 & 86 \\
$3.0\times10^{-4}$ & mild erosion & 0.305 & 0.019 / 0.014 & 5 \\
$1.0\times10^{-4}$ & collapsed & 0.117 & 0.000 / 0.000 & 0 \\
\bottomrule
\end{tabular}
\end{table}

\begin{figure}[h]
\centering
\includegraphics[width=\linewidth]{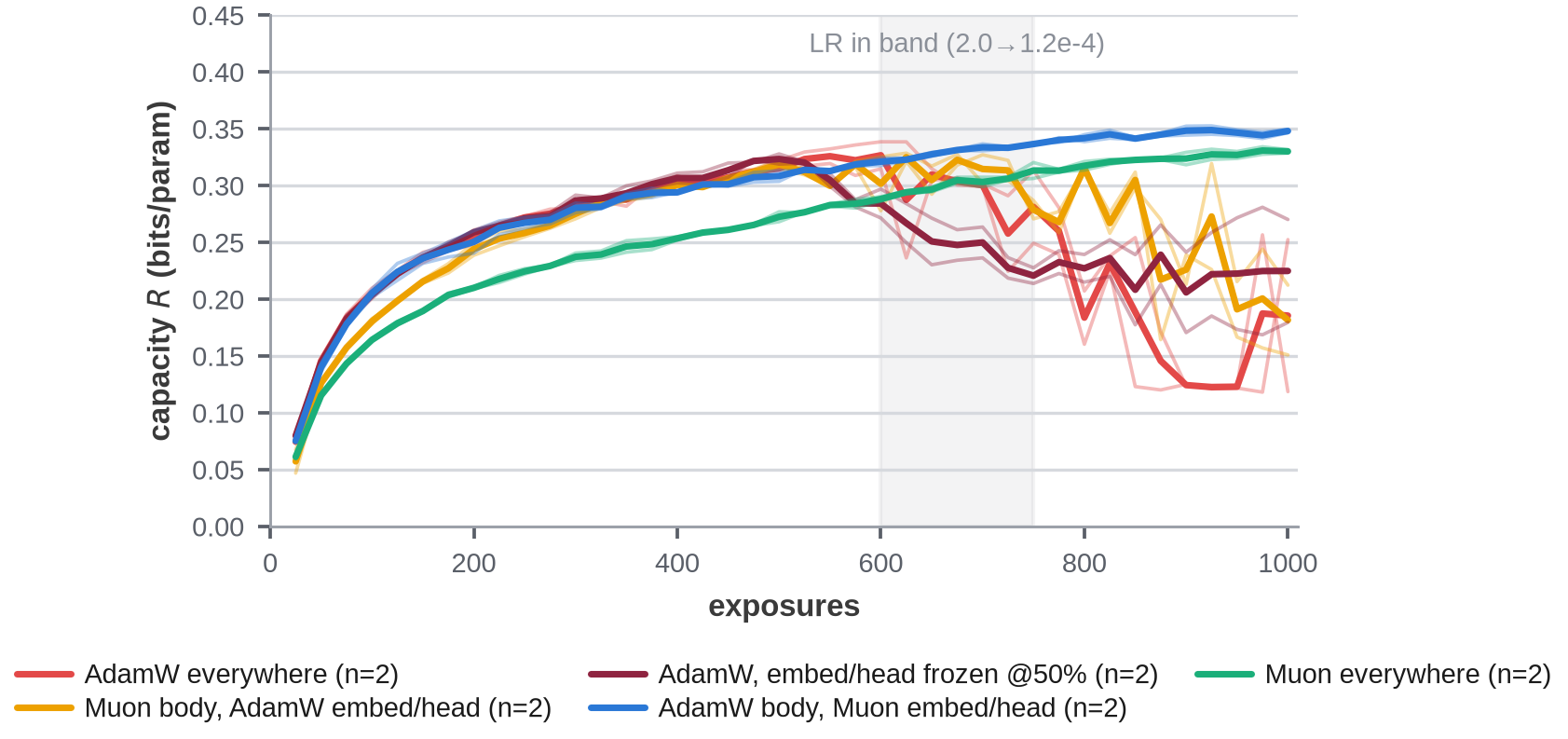}
\caption{Capacity traces for the optimizer-placement runs of Table~\ref{tab:opt} under cosine (thin: seeds; bold: mean).}
\label{fig:placement}
\end{figure}

\section{PTQ Method Details}
\label{app:ptq}

We implement GPTQ as block-sequential error compensation on our exact group-64 lattice (calibration Hessian, standard damping, Cholesky update, projection onto our integer or ternary lattice). At int8, it lands within $0.05$ of the fp16 reference on our evaluator, which we can still reproduce from the archived sweep; the layer-wise reconstruction-error measurement we ran during development was not archived. AWQ follows the standard channel-scaling formulation with a per-layer grid search over the scaling exponent; NF4 uses the 16-level NormalFloat codebook at group-64. A calibration-source sensitivity check (held-out biographies, training text, random tokens) moved retention by less than $0.03$ with the method ranking unchanged.

\begin{figure}[h]
\centering
\includegraphics[width=0.72\linewidth]{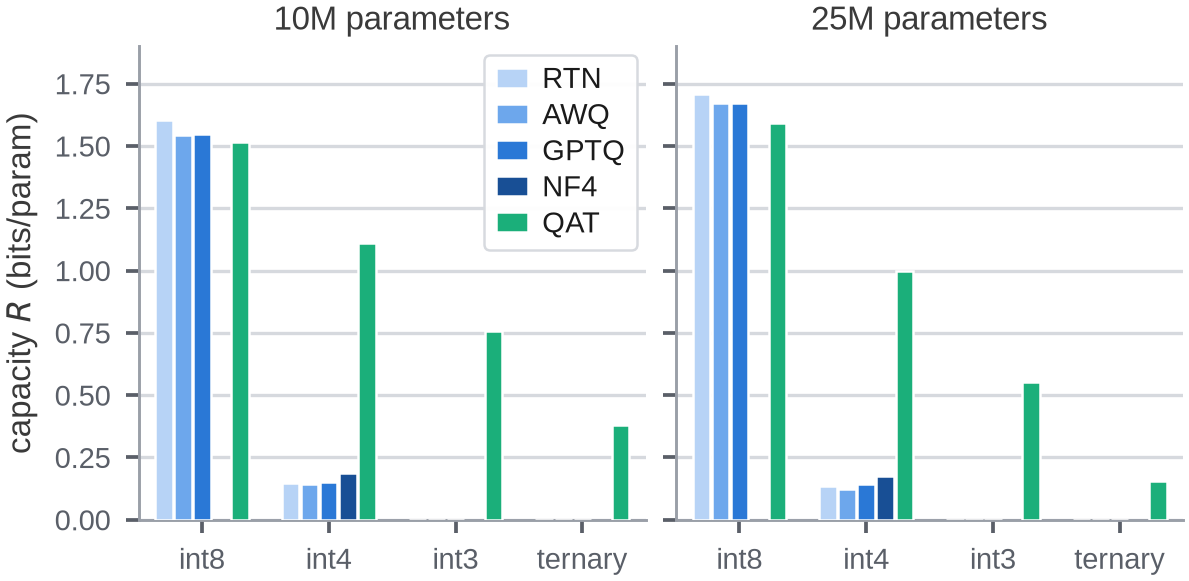}
\caption{Capacity by method and bit-width at 10M and 25M: QAT (green) against the PTQ methods (blue family). PTQ curves quantize seed 0 of each fp16 cell, so their int8 values track that seed; the QAT curve is the cell mean, as in Table~\ref{tab:ptq}.}
\label{fig:ptq}
\end{figure}

\section{Morris-Style Second Capacity Estimator}
\label{app:morris}

The compression-gain estimator of \citet{morris2025memorization} (the difference in per-token compression between held-out and trained-on people, requiring no knowledge-token selection and no clamp) agrees with our primary metric at 10M (retention $0.235$ versus $0.200$) and sits above both clamp conventions at 25M ($0.185$, against $0.090$ and $0.126$), consistent with the aggregate clamp undercounting a collapsed model's residual knowledge (Appendix~\ref{app:convention}). fp16 measures $3.0$--$3.3$ bits per parameter under it at 10M and 25M, and $2.5$--$2.7$ at 50M, below the $3.5$--$3.8$ range they report. Template-level knowledge is precision-insensitive at $9.4$--$9.9$ bits per token across precisions, including ternary: the collapse is specific to facts about individuals.

Scoring every checkpoint both ways, the two estimators agree up to a constant on pre-collapse models (Morris over ours is $1.53$--$1.71\times$ for pre-collapse ternary and $1.55$--$2.13\times$ for fp16) and diverge only on collapsed ones ($3.6$--$5.3\times$). Because the estimator subtracts held-out people drawn from the same generator, template and grammar cancel, so the excess is person-specific information outside the six scored attributes; which estimator better captures retained factual knowledge cannot be settled without a token-level decomposition we have not run. Two cautions follow. It does not register the 25M collapse at all: collapsed cosine scores $0.567$ against the pre-collapse constant-LR checkpoint's $0.557$, so the $1.2\times$ WSD-over-cosine ratio it returns at 25M (Table~\ref{tab:conv}) is measured on a quantity on which the effect is nearly absent. And our implementation caps the held-out cross-entropy at $\log V$, since a trained model is anti-predictive on unseen people; that cap is what produces the negative 50M cell in Table~\ref{tab:conv} (uncapped, $+0.643$ against its sibling's $+0.706$), so the entry is an artifact of the choice rather than a property of the checkpoint.

\section{Attribute-Level Structure of the Collapse}
\label{app:attributes}

\begin{figure}[h]
\centering
\includegraphics[width=\linewidth]{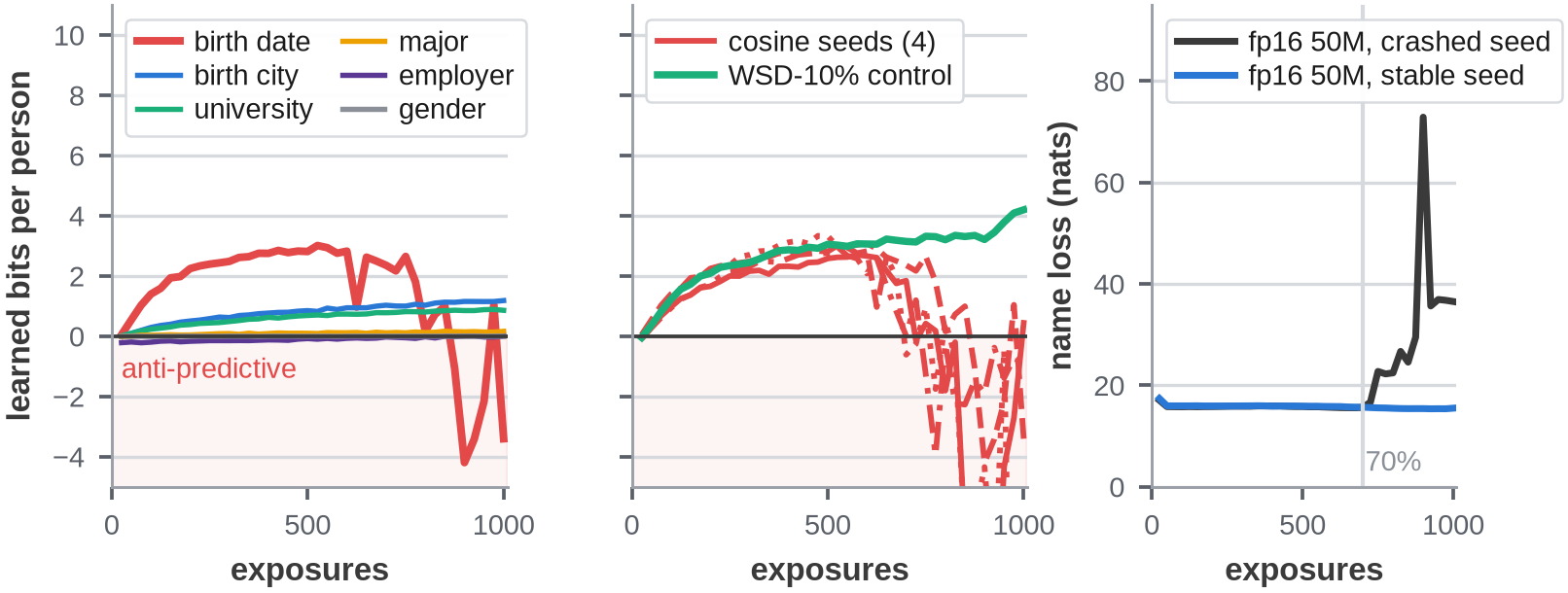}
\caption{Attribute-level view of the collapse. \textbf{Left}: per-attribute learned bits per person in one 25M ternary cosine run. \textbf{Middle}: birth date across four cosine seeds and a WSD control. \textbf{Right}: name-prediction loss of the crashed 50M fp16 seed.}
\label{fig:bistab}
\end{figure}

Pooling every post-peak checkpoint (89 to 179 samples across nine crashing full-load ternary runs) and marking an attribute as surviving if it sits at or above half its within-run peak, birth date survives only 25--33\% of post-peak checkpoints, while birth city, university, and major survive essentially all. Employer and gender, which carry under a bit per person, survive at the same low rate as birth date (both 24\%), a noisy ratio on a near-zero quantity; survival does not order by cardinality. The WSD-33\% seed-0 final ($0.259$) is a real late capture, not an evaluation artifact: its subsample trace fell from $0.365$ to $0.263$ between exposures 975 and 1000, at a learning rate near $5\times10^{-5}$. At 50M, the name behaves as birth date does at 25M: the crashed fp16 seed and the int8 seed both end with negative learned bits there (Appendix~\ref{app:audit}), and the name is the only other attribute whose loss has an irreducible floor at $\ln N$.

\section{Capacity-Landscape Probe}
\label{app:landscape}

\begin{figure}[h]
\centering
\includegraphics[width=\linewidth]{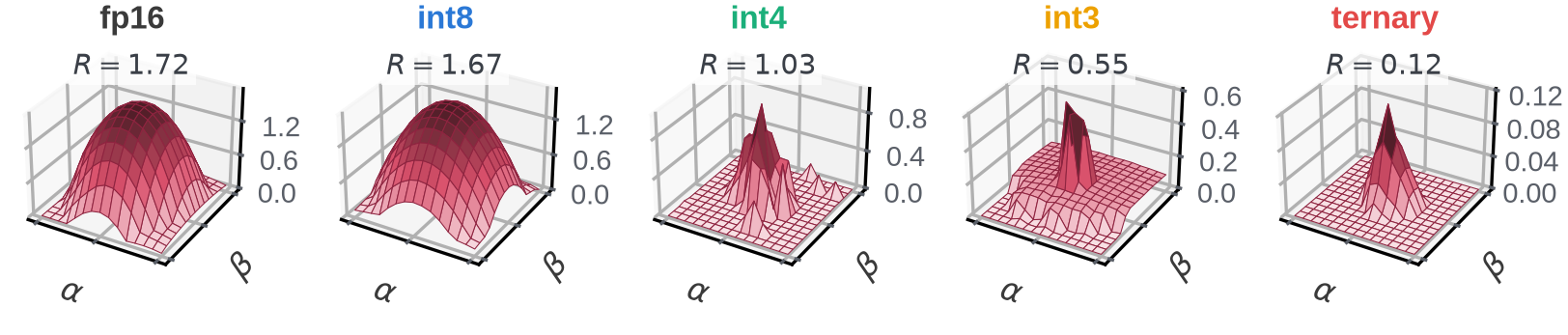}
\caption{Capacity $R$ re-measured while perturbing each trained 25M checkpoint over the same parameter-space range ($\alpha,\beta\in[-0.04,0.04]$, filter-normalized, on every panel). The ternary curve is centered on the two \texttt{grid\_25m} seeds, not the pooled six-seed cell.}
\label{fig:landscape}
\end{figure}

\begin{figure}[h]
\centering
\includegraphics[width=\linewidth]{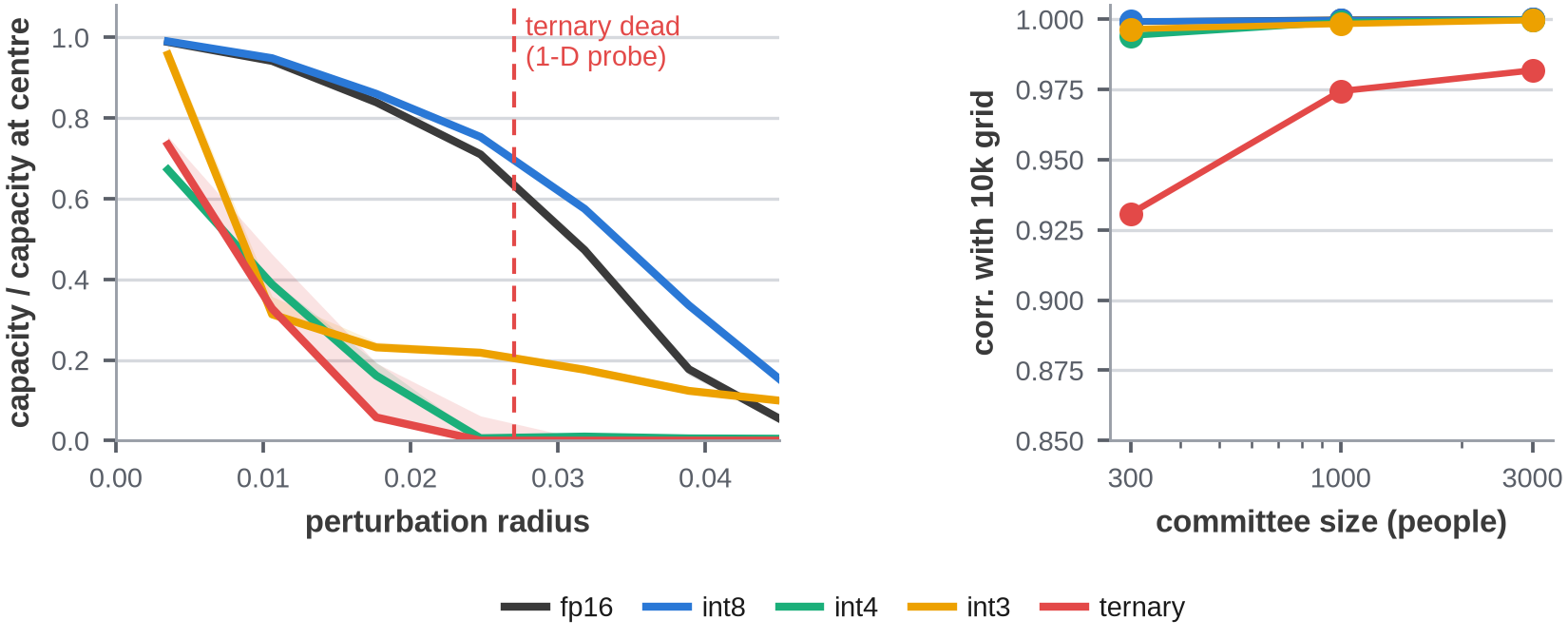}
\caption{One-dimensional capacity basin profiles through trained 25M checkpoints at each precision.}
\label{fig:basin}
\end{figure}

We adapt the filter-normalized random-direction technique of \citet{li2018visualizing} with capacity on the vertical axis. For each trained 25M seed-0 checkpoint, we draw two Gaussian directions, normalize each output row to the norm of the corresponding weight row, and evaluate $R$ on a $15\times15$ grid of offsets $\theta+\alpha d_1+\beta d_2$, $\alpha,\beta\in[-0.04,0.04]$, using a fixed committee of 10,000 individuals. Along $\beta=0$, the ternary surface loses about 30\% of its central capacity within one grid step ($|\alpha|=0.006$) and reaches exactly zero by $|\alpha|\approx0.03$; the grid spacing of $0.006$ is the resolution of this statement. Committees of 300, 1,000, and 3,000 reproduce every surface at Pearson $r\geq0.994$ for fp16 through int3 and $r\geq0.93$ for ternary; the ternary center value ($R=0.119$) agrees within 3\% with the full-population evaluation ($0.122$). Figure~\ref{fig:basin} shows one-dimensional profiles through the same checkpoints. The probe is static and correlational; it is consistent with the dynamical account of \S\ref{sec:mech} but does not test it.

\end{document}